\documentclass[sigconf,authorversion]{acmart}

\usepackage{graphicx}
\usepackage{amsmath}
\usepackage{mathtools}
\usepackage{amsfonts}
\usepackage{bm}
\usepackage{booktabs}
\usepackage{bbm}
\usepackage{multirow}
\usepackage{enumitem}
\usepackage{wrapfig}
\usepackage[tableposition=bottom]{caption}
\usepackage{subcaption}
\usepackage{float}
\usepackage{mathrsfs}
\usepackage{color}
\usepackage{arydshln}
\usepackage{mathtools}
\usepackage[linesnumbered,boxed,ruled,commentsnumbered]{algorithm2e}
\usepackage[utf8]{inputenc}
\usepackage[T1]{fontenc}
\usepackage{circledsteps}
\pgfkeys{/csteps/inner color=white}
\pgfkeys{/csteps/outer color=white}
\pgfkeys{/csteps/fill color=black}
\pgfkeys{/csteps/inner ysep=3pt}
\pgfkeys{/csteps/inner xsep=3pt}

\usepackage{url}
\usepackage{nicefrac}
\usepackage{microtype}
\usepackage{xcolor}
\usepackage{pifont}
\usepackage{hyperref}
\hypersetup{
     colorlinks=true,
     linkcolor=blue,
     filecolor=blue,
     citecolor = black,      
     urlcolor=cyan,
     }

\theoremstyle{definition}

\newtheorem{theorem}{Theorem}
\newtheorem{corollary}{Corollary}
\newtheorem{definition}{Definition}

\setenumerate[1]{itemsep=0pt,partopsep=0pt,parsep=\parskip,topsep=5pt}
\setitemize[1]{itemsep=0pt,partopsep=0pt,parsep=\parskip,topsep=5pt}
\setdescription{itemsep=0pt,partopsep=0pt,parsep=\parskip,topsep=5pt}

\newcommand{\ms}[2]{{#1\tiny{$\pm$#2}}}

\newcommand{\ums}[2]{{\underline{#1}\tiny{$\pm$#2}}}
\newcommand{\ubms}[2]{{\underline{\textbf{#1}}\tiny{$\pm$#2}}}

\newcommand{\reaarr}{\vcenter{\hbox{\includegraphics[scale=0.3]{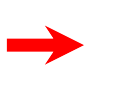}}}}

\usepackage{etoolbox}

\AtBeginDocument{%
  \providecommand\BibTeX{{%
    \normalfont B\kern-0.5em{\scshape i\kern-0.25em b}\kern-0.8em\TeX}}}

\copyrightyear{2022}
\acmYear{2022}
\setcopyright{acmcopyright}\acmConference[WWW '22]{Proceedings of the ACM Web Conference 2022}{April 25--29, 2022}{Virtual Event, Lyon, France}
\acmBooktitle{Proceedings of the ACM Web Conference 2022 (WWW '22), April 25--29, 2022, Virtual Event, Lyon, France}
\acmPrice{15.00}
\acmDOI{10.1145/3485447.3512170}
\acmISBN{978-1-4503-9096-5/22/04}

\begin{document}

\title[Geometric Graph Representation Learning]{Geometric Graph Representation Learning via \texorpdfstring{\\}{} Maximizing Rate Reduction}


\author{Xiaotian Han}
\orcid{0000-0002-1344-3658}
\affiliation{%
  \institution{Texas A\&M University, TX, USA}
  \country{}
}
\email{han@tamu.edu}

\author{Zhimeng Jiang}
\affiliation{%
  \institution{Texas A\&M University, TX, USA}
  \country{}
}
\email{zhimengj@tamu.edu}

\author{Ninghao Liu}
\affiliation{%
  \institution{University of Georgia, GA, USA}
  \country{}
}
\email{ninghao.liu@uga.edu}

\author{Qingquan Song}
\affiliation{%
  \institution{LinkedIn, CA, USA}
  \country{}
}
\email{qsong@linkedin.com}

\author{Jundong Li}
\affiliation{%
  \institution{University of Virginia, VA, USA}
  \country{}
  }
\email{jundong@virginia.edu }

\author{Xia Hu}
\affiliation{%
  \institution{Rice University, TX, USA}
  \country{}
  }
\email{xia.hu@rice.edu}


\begin{abstract}
Learning discriminative node representations benefits various downstream tasks in graph analysis such as community detection and node classification. Existing graph representation learning methods (e.g., based on random walk and contrastive learning) are limited to maximizing the \emph{local} similarity of connected nodes. Such pair-wise learning schemes could fail to capture the \emph{global} distribution of representations, since it has no explicit constraints on the global geometric properties of representation space. To this end, we propose \underline{G}eometric \underline{G}raph \underline{R}epresentation Learning ($\mathrm{G}^2\mathrm{R}$) to learn node representations in an unsupervised manner via maximizing rate reduction. In this way, $\mathrm{G}^2\mathrm{R}$ maps nodes in distinct groups (implicitly stored in the adjacency matrix) into different subspaces, while each subspace is compact and different subspaces are dispersedly distributed. $\mathrm{G}^2\mathrm{R}$ adopts a graph neural network as the encoder and maximizes the rate reduction with the adjacency matrix. Furthermore, we theoretically and empirically demonstrate that rate reduction maximization is equivalent to maximizing the principal angles between different subspaces. Experiments on real-world datasets show that $\mathrm{G}^2\mathrm{R}$ outperforms various baselines on node classification and community detection tasks.
\end{abstract}

\ccsdesc[500]{Computing methodologies~Unsupervised learning}
\ccsdesc[500]{Computing methodologies~Neural networks}
\ccsdesc[500]{Networks~Social media networks}

\keywords{Graph representation learning, Graph neural networks, Unsupervised learning, Rate reduction}
\maketitle

\section{Introduction}\label{sec:Introduction}
Learning effective node representations~\cite{hamilton2017inductive} benefits various graph analytical tasks, such as social science~\cite{ying2018graph}, chemistry~\cite{de2018molgan}, and biology~\cite{zitnik2017predicting}. 
Recently, graph neural networks (GNNs)~\cite{zhou2020graph,wu2020comprehensive} have become dominant technique to process graph-structured data, which typically need high-quality labels as supervision. However, acquiring labels for graphs could be time-consuming and unaffordable. The noise in labels will also negatively affect model training, thus limiting the performance of GNNs. In this regard, learning high-quality low-dimensional representations with GNNs in an unsupervised manner is essential for many downstream tasks.

Recently, many research efforts have been devoted to learning node representations in an unsupervised manner. Most existing methods can be divided into two categories, including random walk based methods~\cite{perozzi2014deepwalk,grover2016node2vec} and contrastive learning methods~\cite{you2020graph, velivckovic2018deep}. 
These methods learn node representations mainly through controlling the representation similarity of connected nodes. For example, DeepWalk~\cite{perozzi2014deepwalk} considers the similarity of nodes in the same context window of random walks. GRACE~\cite{zhu2020deep} uses contrastive learning to model the similarity of connected nodes with features. Such a pair-wise learning scheme encourages the \emph{local} representation similarity between connected nodes, but could fail to capture the \emph{global} distribution of node representations, since it does not directly specify the geometrical property of latent space.

To bridge the gap, we propose to explicitly control the global geometrical discriminativeness of node representations instead of only enforce the local similarity of connected nodes. However, directly constraining the global geometric property of the representation space remains challenging due to the following reasons. 
First, it is difficult to measure the diversity of representations within the same group or across different groups, since the global information such as community distribution is not available in unsupervised settings. Pre-computed node clustering will not fully solve the problem, because there is no guarantee on the quality of resultant clusters, and it even introduces noisy supervised information. Second, it is hard to balance the global geometric property and local similarity, especially when considering the downstream tasks. Since the local similarity of connected nodes is crucial to the performance of downstream tasks, we need to control the global geometric property and local similarity simultaneously.

\begin{figure}[t]
    \includegraphics[width=0.48\textwidth]{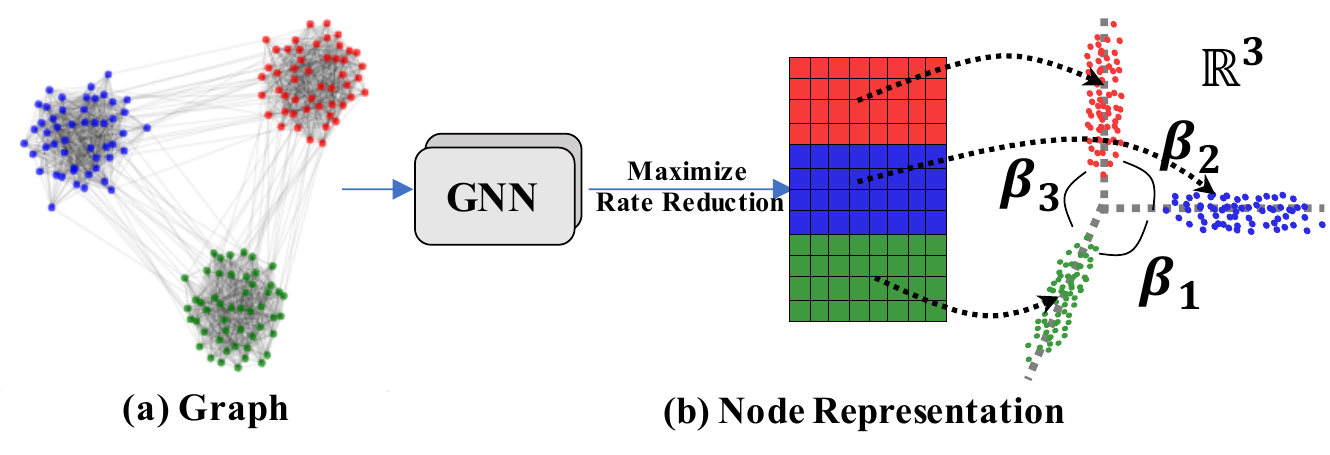}
    \vspace{-20pt}
    \caption{Overview of $\mathrm{G}^2\mathrm{R}$. It maps nodes in distinct groups (implicitly preserved in adjacency matrix) into different subspaces, while each subspace is compact and different subspaces are dispersedly distributed. Different colors indicate different subspaces and node groups.}\label{fig:intro}
     \vspace{-10pt}
\end{figure}

To address the above challenges, we propose \underline{G}eometric \underline{G}raph \underline{R}epresentation Learning ($\mathrm{G}^2\mathrm{R}$) to learn node representations via maximizing coding rate reduction. First, we leverage the coding rate~\cite{yu2020learning} to estimate the diversity of a set of node representations. A higher coding rate means representations are diversely spread in the latent space. Also, we define \textit{rate reduction} as the difference of coding rates between representations of the entire nodes and each of the groups. Then, we maximize the rate reduction to learn geometrically discriminative representations. A higher rate reduction means node representations are close to each other within each group, while they are far away from each other across different groups. This can be achieved even without explicitly knowing the node-group assignments. We use graph neural networks as the encoder to generate node representations, and map the nodes in the same group into the identical latent subspace. Specifically, Figure~\ref{fig:intro} presents an intuitive overview of $\mathrm{G}^2\mathrm{R}$. The nodes in green, blue and red (Figure~\ref{fig:intro}(a)) are projected to different subspaces (Figure~\ref{fig:intro}(b)), and the difference between subspaces are maximized. The main contributions are summarized as follows:
\begin{itemize}[leftmargin=0.4cm]
\item We propose a new objective for unsupervised graph learning via maximizing rate reduction, which encourages the encoder to learn discriminative node representations with only the adjacency matrix (Section~\ref{sec:meth}).

\item We provide theoretical justification for the proposed method from the perspective of maximizing the principal angles between different latent subspaces. (Section~\ref{sec:theo}).

\item Experimental results on synthetic graphs validate the theoretical results of the proposed method~(Section~\ref{sec:exp:syn}). And the visualization results on real-world datasets also verify that resultant node representations are nearly orthogonal~(Section~\ref{sec:exp:vis}).

\item Experimental results on real-world datasets show that the node representations learned by $\mathrm{G}^2\mathrm{R}$ are consistently competitive on the node classification and community detection tasks. Besides, $\mathrm{G}^2\mathrm{R}$ achieves comparable performance to supervised baselines on node classification~(Section~\ref{sec:exp:unsuper} and ~\ref{sec:exp:super}).
\end{itemize}

\section{Preliminaries}\label{sec:pre}
In this section, we present essential preliminaries. First, we introduce the notations in this work. Then we introduce the idea of rate reduction for representation learning.

\subsection{Notations}\label{sec:notation}
A graph is denoted as $\mathcal{G} = \{ \mathcal{V}, \mathcal{E} \}$, where $\mathcal{V}$ is the node set and $\mathcal{E}$ is the edge set. The number of nodes is $N$. The adjacency matrix is denoted as $\mathbf{A} = [\mathbf{a}_1,\mathbf{a}_2, \cdots, \mathbf{a}_N ]\in \mathbb{R}^{N \times N}$, where $\mathbf{a}_i$ is the neighbor indicator vector of node $i$. The feature matrix is $\mathbf{X} = [\mathbf{x}_1,\mathbf{x}_2, \cdots, \mathbf{x}_N ] \in \mathbb{R}^{d_0 \times N}$, where $d_0$ is the dimension of node features. A graph neural network encoder is denoted as $ \mathrm{Enc}( \mathbf{A}, \mathbf{X} )$, which transforms the nodes to representations $\mathbf{Z} = [\mathbf{z}_1,\mathbf{z}_2,\cdots,\mathbf{z}_N]\in \mathbb{R}^{d \times N}$, where $d$ is the dimension of $\mathbf{z}_i$.

\subsection{Representation Learning via Maximizing Rate Reduction}
In this part, we introduce rate reduction~\cite{yu2020learning}, which was proposed to learn diverse and discriminative representations. The coding rate~\cite{ma2007segmentation} is a metric in information theory to measure the compactness of representations over all data instances. A lower coding rate means more compact representations.
Suppose a set of instances can be divided into multiple non-overlapping groups. Rate reduction measures the difference of coding rates between the entire dataset and the sum of that of all groups. Higher rate reduction implies more discriminative representation among different groups and more compact representation within the same group.

\noindent\textbf{Representation Compactness for the Entire Dataset}. Let $f(\cdot)$ denote the encoder, where the representation of a data instance $\mathbf{x}_i$ is $\mathbf{z}_i = f(\mathbf{x}_i) \in\mathbb{R}^{d}$.
Given the representations $\mathbf{Z} = [\mathbf{z}_1,\mathbf{z}_2,\cdots,\mathbf{z}_N]\in\mathbb{R}^{d\times N}$ of all data instances, the coding rate is defined as the number of binary bits to encode $\mathbf{Z}$, which is estimated as below \cite{ma2007segmentation}:
\begin{equation}\label{equ:Rz}
\begin{aligned}
R(\mathbf{Z}, \epsilon) \doteq \frac{1}{2} \log \operatorname{det}\left(\mathbf{I}+\frac{d}{N \epsilon^{2}} \mathbf{Z} \mathbf{Z}^{\top}\right),
\end{aligned}
\end{equation}
where $\mathbf{I}$ is the identity matrix, $N$ and $d$ denote the length and dimension of learned representation $\mathbf{Z}$, and $\epsilon$ is the tolerated reconstruction error (usually set as a heuristic value $0.05$).

\noindent\textbf{Representation Compactness for Groups}. Given $\mathbf{Z} = [\mathbf{z}_1,\mathbf{z}_2,$ $\cdots,\mathbf{z}_N]\in\mathbb{R}^{d\times N}$, we assume the representations can be partitioned to $K$ groups with a probability matrix $\mathbf{\pi}\in \mathbb{R}^{N\times K} $. Here $\pi_{ik} \in [0,1]$ indicates the probability of instance $\mathbf{x}_i$ assigned to the subset $k$, and $\sum_{k=1}^{K} \pi_{ik} = 1$ for any $i\in[N]$. We define the membership matrix for subset $k$ as $\mathbf{\Pi}_k =diag[ \pi_{1k}, \pi_{2k}, \cdots, \pi_{Nk} ] \in \mathbb{R}^{N\times N} $, and the membership matrices for all groups are denoted as $\mathbf{\Pi} = \{ \mathbf{\Pi}_k |k=[K]\}$. Thus, the coding rate for the entire dataset is equal to the summation of coding rate for each subset:
\begin{equation}\label{equ:Rcz}
    R^{c}(\mathbf{Z}, \epsilon|\mathbf{\Pi}) \doteq \sum_{k=1}^{K}\frac{ tr( \mathbf{\Pi}_k ) }{2\cdot N}
    \cdot\log \operatorname{det}\left(\mathbf{I}+\frac{d}{ tr( \mathbf{\Pi}_k ) \epsilon^{2}} \mathbf{Z} \mathbf{\Pi}_k \mathbf{Z}^{\top}\right).
\end{equation}

\noindent\textbf{Rate Reduction for Representation Learning}. Intuitively, the learned representations should be diverse in order to distinguish instances from different groups. That is, \emph{i) the coding rate for the entire dataset should be as large as possible to encourage diverse representations}
; \emph{ii) the representations for different groups should span different subspaces and be compacted within a small volume for each subspace.} 
Therefore, a good representation achieves a \emph{larger} rate reduction (i.e., difference between the coding rate for datasets and the summation of that for all groups):
\begin{align}\label{equ:mcrr}
\Delta R(\mathbf{Z}, \mathbf{\Pi}, \epsilon) = R(\mathbf{Z}, \epsilon) - R^{c}(\mathbf{Z}, \epsilon|\mathbf{\Pi}).
\end{align}
Note that the rate reduction is monotonic with respect to the norm of representation $\mathbf{Z}$. So we need to normalize the scale of the learned features, each $\mathbf{z}_i$ in $\mathbf{Z}$ is normalized in our case.

\section{Methodology}\label{sec:meth}

In this section, we introduce our $\mathrm{G}^2\mathrm{R}$ model based on rate reduction for unsupervised graph representation learning. Specifically, we first introduce how to compute the coding rate of node representations for the nodes in the whole graph and in each group, respectively. Then, we introduce how to incorporate rate reduction into the design of the learning objective and how to train $\mathrm{G}^2\mathrm{R}$. 

\subsection{Coding Rate of Node Representations}
Our goal is to learn an \emph{encoder} $\mathbf{Z} = \mathrm{Enc}( \mathbf{A}, \mathbf{X}| \theta )$, which transforms the graph to the node representations, where $\mathbf{Z}\in \mathbb{R}^{d \times N}$ and $\theta$ is the encoder parameters to be optimized. The encoder in this work is instantiated as a graph neural network. The learned node representations will be used for various downstream applications, such as node classification and community detection.

\subsubsection{Computing Coding Rate of Entire Node Representations}
\noindent Let $\mathbf{Z} = [\mathbf{z}_1,\mathbf{z}_2,\cdots,\mathbf{z}_N]\in\mathbb{R}^{d\times N}$ be the node representations. We use coding rate to estimate the number of bits for representing $\mathbf{Z}$ within a specific tolerated reconstruction error $\epsilon$. Therefore, in graph $\mathcal{G}$, the coding rate of node representations is $R_{\mathcal{G}}(\mathbf{Z}, \epsilon) = R(\mathbf{Z}, \epsilon)$ as defined in Equation~\ref{equ:Rz}. A larger $R_{\mathcal{G}}$ corresponds to more diverse representations across nodes, while a smaller $R_{\mathcal{G}}$ means a more compact representation distribution. 

\subsubsection{Computing Coding Rate for Groups} 
To enforce the connected nodes have the similar representations, we cast the \emph{node $i$ and its neighbors as a group} and then map them to identical subspace. To do this, we assemble the membership matrix based on the adjacency matrix. The adjacency matrix is $\mathbf{A} = [ \mathbf{a}_1,\mathbf{a}_2,\cdots, \mathbf{a}_N  ] \in \mathbb{R}^{N\times N}$ where $\mathbf{a}_i \in \mathbb{R}^{N}$ is the neighbor indicator vector of node $i$.
Then we assign membership matrix for the node group as $\mathbf{A}_i= diag( \mathbf{a}_i ) \in \mathbb{R}^{N\times N}$. The coding rate for the group of node representations with membership matrix $\mathbf{A}_i$ is as follows:
\begin{equation}\label{equ:gRcz}
    R^{c}_{\mathcal{G}}(\mathbf{Z}, \epsilon|\mathbf{A}_i) \doteq \frac{ \mathrm{tr}( \mathbf{A}_i ) }{2N}
    \cdot\log \operatorname{det}\left(\mathbf{I}+\frac{d}{ \mathrm{tr}( \mathbf{A}_i ) \epsilon^{2}} \mathbf{Z} \mathbf{A}_i \mathbf{Z}^{\top}\right).
\end{equation}
Thus for all nodes in the graph, the membership matrix set will be $\mathcal{A}=\{ \mathbf{A}_i\in \mathbb{R}^{N \times N}, i\in [N] \}$. Since the $\sum_{i=1}^{N}\mathbf{A}_i = \mathbf{D}$, where $\mathbf{D}={diag}( d_1, d_2, \cdots, d_N ) \in \mathbb{R}^{N\times N}$ is degree matrix and $d_i$ is the degree of node $i$. The different groups of node is overlapping and will be computed multiple times, thus we normalize the coding rate of node representations for groups with the average degree $\bar{d}$ of all nodes. Consequently, the sum of the coding rate of node representations for each group is given as the following:
\begin{equation}\label{equ:gallRcz}
    R^{c}_{\mathcal{G}}(\mathbf{Z}, \epsilon|\mathcal{A}) \doteq \frac{1}{ \bar{d} }\sum_{i=1}^{N}\frac{ \mathrm{tr}( \mathbf{A}_i ) }{2N}
    \cdot\log \operatorname{det}\left(\mathbf{I}+\frac{d}{ \mathrm{tr}( \mathbf{A}_i ) \epsilon^{2}} \mathbf{Z} \mathbf{A}_i \mathbf{Z}^{\top}\right),
\end{equation}
where $N$ is the total number of nodes in the graph, $\bar{d}$ is the average degree of nodes, and $\mathcal{A}$ is the membership matrix set.

\subsection{Rate Reduction Maximization for Training}

\subsubsection{Objective function}
Combining Equations~\eqref{equ:gRcz} and~\eqref{equ:gallRcz}, the rate reduction for the graph with adjacency matrix $\mathbf{A}$ is given as follows:
\begin{equation}\label{equ:loss}
\begin{split}
    \Delta R_{\mathcal{G}}(\mathbf{Z}, \mathbf{A}, \epsilon) &= R_{\mathcal{G}}(\mathbf{Z}, \epsilon) - R^{c}_{\mathcal{G}}(\mathbf{Z}, \epsilon \mid \mathcal{A} )\\
    & \doteq \frac{1}{2} \log \operatorname{det}\left(\mathbf{I}+\frac{d}{N \epsilon^{2}} \mathbf{Z} \mathbf{Z}^{\top}\right)\\
    &\quad-\frac{1}{\bar{d}}\sum_{i=1}^{N}\frac{ \mathrm{tr}( \mathbf{A}_i ) }{2N}
    \cdot\log \operatorname{det}\left(\mathbf{I}+\frac{d}{ \mathrm{tr}( \mathbf{A}_i ) \epsilon^{2}} \mathbf{Z} \mathbf{A}_i \mathbf{Z}^{\top}\right).
\end{split}
\end{equation}
In practice, we control the strength of compactness of the node representations by adding two hyperparameters $\gamma_1$ and $\gamma_2$ to the first term in Equation~\eqref{equ:loss}. The $\gamma_1$ controls compression of the node representations while the $\gamma_2$ balances the coding rate of the entire node representations and that of the groups. Thus we have
\begin{equation}\label{equ:gamloss}
\begin{split}
    &\Delta R_{\mathcal{G}}(\mathbf{Z}, \mathbf{A}, \epsilon, \gamma_1, \gamma_2) \\
    &\quad\doteq \frac{1}{2\gamma_1} \log \operatorname{det}\left(\mathbf{I}+\frac{d\gamma_2}{N \epsilon^{2}} \mathbf{Z} \mathbf{Z}^{\top}\right)\\
    &\quad\quad-\frac{1}{ \bar{d} }\sum_{i=1}^{N}\frac{ \mathrm{tr}( \mathbf{A}_i ) }{2N}
    \cdot\log \operatorname{det}\left(\mathbf{I}+\frac{d}{ \mathrm{tr}( \mathbf{A}_i ) \epsilon^{2}} \mathbf{Z} \mathbf{A}_i \mathbf{Z}^{\top}\right),
\end{split}
\end{equation}
where $\epsilon$, $\gamma_1$, and $\gamma_2$ serve as the hyperparameters of our model. 

\subsubsection{Model Training}
We adopt graph neural network as the encoder to transform the input graph to node representations, where $\mathbf{Z}=\text{GNN}( \mathbf{X}, \mathbf{A}|\theta)$ and $\theta$ denotes the parameters to be optimized. The output of the last GNN layer is the learned node representations, which is $L-1$ normalized as mentioned before. The parameters $\theta$ will be optimized by maximizing the following objective:
\begin{equation}\label{equ:obj}
    \begin{split}
        \max\limits_{\theta} \Delta R_{\mathcal{G}}( \text{GNN}(\mathbf{X},\mathbf{A}|\theta),\mathbf{A}, \epsilon, \gamma_1, \gamma_2),
    \end{split}
\end{equation}
where $\epsilon$, $\gamma_1$, and $\gamma_2$ serve as the hyperparameters of our model. We also conduct experiments to explore the effect of hyperparameters $\gamma_1$ and $\gamma_2$ in Section~\ref{exp:hp}. We set hyperparameters $\epsilon$ to a heuristic value $0.05$. For large graphs, the adjacency matrix is large and the length of membership matrix set is $|\mathcal{A}|=N$, thus we need to compute coding rate for groups $N$ times in Equations~\eqref{equ:gallRcz} and \eqref{equ:loss}. To reduce the computational complexity, we randomly sample fixed number $N_s$ rows of adjacency matrix for each training batch. Then we use sampled adjacency matrix to assemble the membership matrix set, which only has $N_s$ membership metrics. Thus we only compute the coding rate $N_s$ times.

\subsubsection{Computational Complexity}
Due to the commutative property~\footnote{ Commutative property of  coding rate: $R(\mathbf{Z},\epsilon) \doteq \frac{1}{2}\log \det\left(\mathbf{I} + \frac{d}{m\epsilon^{2}}\mathbf{Z}\mathbf{Z}^{\top}\right) =\frac{1}{2}\log\det\left(\mathbf{I} + \frac{d}{m\epsilon^{2}}\mathbf{Z}^{\top}\mathbf{Z}\right)$} of coding rate, computational complexity of the proposal is not high. In this work, we have $\mathbf{Z}\in \mathbb{R}^{N\times d}$, where $d$ is the dimension of node representations and $N$ is the total number of nodes. So we have $\mathbf{Z}^{\top}\mathbf{Z}\in \mathbb{R}^{d\times d}$ and $\mathbf{Z}\mathbf{Z}^{\top}\in \mathbb{R}^{N\times N}$.  Even though the computation of $\log \det\left(\mathbf{I} + \frac{d}{m\epsilon^{2}}\mathbf{Z}\mathbf{Z}^{\top}\right)$ takes $\mathcal{O}(N^3)$ times, we can compute  $\log\det\left(\mathbf{I} + \frac{d}{m\epsilon^{2}}\mathbf{Z}^{\top}\mathbf{Z}\right)$ instead, which takes $\mathcal{O}(d^3)$ times and $d \ll N$. In our experiment setting, we set $d$ to $512$. Thus the operation $\mathrm{logdet}(\cdot)$ will only take $\mathcal{O}(d^3)$ times, which is constant time and does not depend on the nodes number $N$. Besides, since $\mathbf{Z}\mathbf{Z}^{\top} \in \mathbb{R}^{d\times d}$, the memory usage will not increase while the number of nodes ($N$) increases, leading to the scalability of $\mathrm{G}^2\mathrm{R}$. 

\subsection{Discussion: what is \texorpdfstring{$\mathrm{G}^2\mathrm{R}$}{} doing intuitively?}
To understand the proposed objective function in Equation~\eqref{equ:loss}, we informally discuss the intuition behind it.

\begin{itemize}[leftmargin=0.4cm]
    \item \textbf{The first term enforces diverse node representations space.} Maximizing the first term in Equation~\eqref{equ:loss} tends to increase the diversity of representation vectors for all nodes, thus leading to a more diverse distribution of node representations.
    \item \textbf{The second term enforces more similar representations for connected nodes.} The second term in Equation~\eqref{equ:loss} measures the compactness of the representation of node groups. Minimizing the second term enforces the similarity of node representations. As a result, the learned representations of connected nodes will cluster together, as shown in Figure~\ref{fig:tsne}.
\end{itemize}

\section{Theoretical Justification}\label{sec:theo}
To gain a deeper insight of $\mathrm{G}^2\mathrm{R}$, we theoretically investigate the Equation~\eqref{equ:loss} on an example graph with two communities as a simplified illustration.  Consequently, we prove that $\mathrm{G}^2\mathrm{R}$ maps representations of nodes in different communities to different subspaces and aim to maximize the principal angle~\footnote{The principal angle measures the difference of subspaces. The higher principal angle indicates more discriminative subspaces.} between different subspaces, thus encouraging them to be (nearly) orthogonal. 

\subsection{Principal Angle Between Subspaces}
To measure the difference between two subspaces, we introduce the \textit{principal angle} $\theta$~\cite{miao1992principal} to generalize the angle between subspaces with arbitrary dimensions. We give the formal definition as follows:

\begin{definition}[Principal angle]\label{def:principalangel}
Given subspace $\mathbf{L}, \mathbf{M} \subseteq \mathbb{R}^n$ with $\text{dim} \mathbf{L} = l \geq \text{dim} \mathbf{M} = m $, there are $m$ principal angles between $\mathbf{L}$ and $\mathbf{M}$ denoted as $0 \leq \theta_1 \leq \theta_2 \leq \cdots \leq  \theta_m \leq \frac{\pi}{2}$ between $\mathbf{L}$ and $\mathbf{M}$ are recursively defined, where
\begin{small}
\begin{equation}
\begin{aligned}
\cos(\theta_i) \coloneqq \min\Bigg\{ \frac{<\mathbf{x}, \mathbf{y}>}{||\mathbf{x}|| ||\mathbf{y}||} \Big| \mathbf{x} \in \mathbf{L}, \mathbf{y} \in \mathbf{M}, 
\mathbf{x}\bot \mathbf{x}_k, \mathbf{y}\bot \mathbf{y}_k, k= 1,\cdots,i-1 \Bigg\}. \notag
\end{aligned}
\end{equation}
\end{small}
\end{definition}
\noindent 
We adopt product of sines of principal angles, denoted as $sin\{\mathbf{L}, \mathbf{M}\} = sin\theta_1 \cdots sin\theta_m\in [0,1]$, to measure the difference between two subspaces. Notably, when two subspaces are orthogonal, the product of principal sines equals $1$.

\subsection{Graph with Two Communities}
Without loss of generality, we analyze the graph with two equal-size communities. We assume each community has $M$ nodes. 
The graph adjacency matrix $\mathbf{A}$ is generated from the Bernoulli distribution of matrix $\mathbf{P} \in \mathbb{R}^{2M\times 2M}$. The matrix $\mathbf{P}$ is defined as follows:
\begin{equation}\label{equ:p_in_out}
p_{i,j} = \left\{  
\begin{array}{ll}  
p^{i}, & \text{if nodes}~i,j \text{ are in the same community;} \\ 
p^{o}, & \text{otherwise},
\end{array}  
\right. 
\end{equation}
where $p_{i,j}$ is the element of matrix $\mathbf{P}$ for $i^{th}$ row and $j^{th}$ column. In other words, the relation between $\mathbf{P}$, $\mathbf{A}$ are shown as follows:

\begin{equation}
\scriptsize
\mathbf{P}=
\begin{bmatrix}
\begin{array}{ccc:ccc}
p^{i} &\cdots& p^{i} & p^{o} &\cdots& p^{o} \\
\vdots & \ddots & \vdots &\vdots & \ddots & \vdots \\
p^{i} &\cdots& p^{i} & p^{o} &\cdots& p^{o} \\ \hdashline
p^{o} &\cdots& p^{o} & p^{i} &\cdots& p^{i} \\
\vdots & \ddots & \vdots &\vdots & \ddots & \vdots \\
p^{o} &\cdots& p^{o} & p^{i} &\cdots& p^{i} \\
\end{array}
\end{bmatrix}
\stackrel{ Bern }{\xRightarrow{\hspace{0.4cm}}}
\mathbf{A} \in \mathbb{R}^{2M\times 2M},
\end{equation}
The $i$-th row of adjacency matrix $\mathbf{A}$ is denoted as $\mathbf{a}_i=[a_{i1},\cdots, a_{iN}] \in \mathbb{R}^{N}$, which is generated from Bernoulli distributions $Bern(\mathbf{P}_{i*})$ independently.
To compute the coding rate in graphs, we rewrite the connectivity probability matrix $\mathbf{P}$ as follows:

\begin{equation}\label{equ:pdecom}
\scriptsize
\begin{split}
\mathbf{P} &=
p^{o}\cdot\mathbbm{1}\mathbbm{1}^{\top}
+ \underbrace{
\begin{bmatrix}
\begin{array}{ccc:ccc}
p^{i}-p^{o} &\cdots& p^{i}-p^{o} & 0 &\cdots& 0 \\
\vdots & \ddots & \vdots &\vdots & \ddots & \vdots \\
p^{i}-p^{o} &\cdots& p^{i}-p^{o} & 0 &\cdots& 0 \\ \hdashline
0 &\cdots& 0 & p^{i}-p^{o} &\cdots& p^{i}-p^{o} \\
\vdots & \ddots & \vdots &\vdots & \ddots & \vdots \\
0 &\cdots& 0 & p^{i}-p^{o} &\cdots& p^{i}-p^{o} \\
\end{array}
\end{bmatrix}}_{\triangleq \mathbf{C}},
\end{split}
\end{equation}
where $\mathbbm{1} \in \mathbb{R}^{N \times 1}$ is an all-ones vector and $\mathbbm{1}\mathbbm{1}^{\top} \in \mathbb{R}^{N \times N}$ is an all-ones matrix.
The first term $p^{o}\cdot\mathbbm{1}\mathbbm{1}^{\top}$ extracts the uniform background factor that is equally applied to all edges. The second term in Equation~\eqref{equ:pdecom}  $\mathbf{C} =[ \mathbf{C}_1, \cdots, \mathbf{C}_M, \mathbf{C}_{M+1}, \cdots, \mathbf{C}_{2M}] \in \mathbb{R}^{2M\times 2M} $ tells the difference of node connections in different communities, so we only focus on the second term in the following analysis. 

\subsection{Coding Rate for Graph with Communities}
Since there are two communities, the membership matrices set is defined as $\mathcal{C} = \{ \mathbf{C}_1, \cdots, \mathbf{C}_M, \mathbf{C}_{M+1}, \cdots, \mathbf{C}_{2M} \}$. Since the $\mathbf{C}_1 = \mathbf{C}_2 = \cdots = \mathbf{C}_M$ and $\mathbf{C}_{M+1} = \mathbf{C}_{M+2} = \cdots = \mathbf{C}_{2M}$, we can rewrite the membership matrix to $\mathcal{C} = \{ \underbrace{ \mathbf{C}^1, \cdots, \mathbf{C}^1}_{ M }, \underbrace{ \mathbf{C}^2, \cdots, \mathbf{C}^2}_{ M } \}$ where $\mathbf{C}^1  = \mathbf{C}_1 = \cdots = \mathbf{C}_M$ and $\mathbf{C}^2 = \mathbf{C}_{M+1} = \cdots = \mathbf{C}_{2M}$.

Thus we soften the Equation~\eqref{equ:gRcz} by replacing $\mathbf{A}_i$ with its $\mathbf{C}_i$,
\begin{equation}\label{equ:cRcz}
\begin{split}
    R^{c}_{\mathcal{G}}(\mathbf{Z}, \epsilon|\mathcal{C}) &\doteq \frac{1}{d}\sum_{i=1}^{2M}\frac{ \mathrm{tr}( \mathbf{C}_i ) }{2N}
    \cdot\log \operatorname{det}\left(\mathbf{I}+\frac{d}{ \mathrm{tr}( \mathbf{C}_i ) \epsilon^{2}} \mathbf{Z} \mathbf{C}_i \mathbf{Z}^{\top}\right)\\
    &\doteq \frac{M}{d}\sum_{i=1}^{2}\frac{ \mathrm{tr}( \mathbf{C}^i ) }{2N}
    \cdot\log \operatorname{det}\left(\mathbf{I}+\frac{d}{ \mathrm{tr}( \mathbf{C}^i ) \epsilon^{2}} \mathbf{Z} \mathbf{C}^i \mathbf{Z}^{\top}\right) .
\end{split}
\end{equation}

\noindent The rate reduction will take
\begin{equation}\label{equ:ssbmloss}
    \begin{split}
    \Delta R_{\mathcal{G}}( \mathbf{Z}, \mathcal{C}, \epsilon ) &= R_{\mathcal{G}}( \mathbf{Z},\epsilon ) -  R^c_{\mathcal{G}}( \mathbf{Z}, \epsilon |\mathcal{C}  )\\
    &=\sum_{j=1}^{2}\text{log}\left( \frac{  \text{det}^{\frac{1}{4}} \left( \mathbf{I} + \frac{d}{N\epsilon^2}\mathbf{Z}_j^{\top}\mathbf{Z}_j \right) }{ \text{det}^{\frac{p^{i} - p^{o}}{2N}} \left( \mathbf{I} + \frac{d }{M\epsilon^2}\mathbf{Z}_j^{\top}\mathbf{Z}_j \right)} \right) + \frac{1}{2}\cdot\text{log}\beta.
    \end{split}
\end{equation}

\noindent where $\mathbf{I} + \frac{d}{N\epsilon^2}\mathbf{Z}^{\top}\mathbf{Z} =\mathbf{ \Tilde{Z} }^{\top} \mathbf{ \Tilde{Z} }$ and $\beta =  \text{sin}\{ R(\mathbf{\Tilde{Z}}_1),R( \mathbf{\Tilde{Z}}_{2}) \}$. The detailed proof is provided in Appendix~\ref{sec:app:proof}.

\subsection{Discussion: what is \texorpdfstring{$\mathrm{G}^2\mathrm{R}$}{} doing theoretically?}\label{sec:theo:doing}
Equation~\eqref{equ:ssbmloss} attempts to optimize the principal angle of different subspaces. Different representation subspaces are more distinguishable if $\beta$ is larger. Thus, maximizing the second term in Equation~\eqref{equ:ssbmloss} promises the following desirable properties: 
\begin{itemize}[leftmargin=0.2cm]
    \item \textbf{Inter-communities.} The representations of nodes in different communities are mutually distinguishable. The node representations of different communities should lie in different subspaces and the principal angle of subspaces are maximized (i.e., nearly pairwise orthogonal), which is verified by experimental results in Figure~\ref{fig:synthetic} and Figure~\ref{fig:orth}.
    \item \textbf{Intra-communities.} The representations of nodes in the same community share the same subspace. So the representations of nodes in the same community should be more similar than nodes in different communities.
\end{itemize}
Based on the above analysis, $\mathrm{G}^2\mathrm{R}$ achieves \emph{geometric} representation learning by constraining the distribution of node representations in different subspaces and encouraging different subspaces to be orthogonal. The \emph{geometric} information considers a broader scope of embedding distribution in latent space.

\begin{table*}[t]
    \centering  
    \fontsize{7.5}{8}\selectfont  
    \vspace{-10pt}
    \caption{Performance comparison to unsupervised methods. The accuracy with standard deviation are based on $5$ runs for all methods. The second column shows the information used by the method, where $\mathbf{X},\mathbf{A}$ denote node features and adjacency matrix, respectively. `OOM' means out of memory while running on NVIDIA RTX3090~(24GB memory). `Public/Random' represents the public/random data split. The best performance among baselines is \underline{underlined}. The best performance is in \underline{\textbf{boldface}.} } 
    \label{tab:unsupervised:random}
    \setlength{\tabcolsep}{3pt}
    \begin{tabular}{lccccccccccccccc}

    \toprule
    \multicolumn{2}{c}{\textbf{Statistic}}&\multicolumn{2}{c}{\textbf{Cora}}&\multicolumn{2}{c}{\textbf{CiteSeer}}&\multicolumn{2}{c}{\textbf{PubMed}} &\multirow{1}{*}{\textbf{CoraFull}}   &\multirow{1}{*}{\textbf{CS}} &\multirow{1}{*}{\textbf{Physics}} & \multirow{1}{*}{\textbf{Computers}} &\multirow{1}{*}{\textbf{Photo} } \\
    \cmidrule(l{0pt}r{5pt}){1-2}\cmidrule(l{5pt}r{5pt}){3-4}\cmidrule(l{5pt}r{5pt}){5-6}\cmidrule(l{5pt}r{5pt}){7-8} \cmidrule(l{5pt}r{5pt}){9-9} \cmidrule(l{5pt}r{5pt}){10-10} \cmidrule(l{5pt}r{5pt}){11-11} \cmidrule(l{5pt}r{5pt}){12-12} \cmidrule(l{5pt}r{5pt}){13-13}
    \multicolumn{2}{c}{\textbf{\#Nodes}}        &\multicolumn{2}{c}{2708}           &\multicolumn{2}{c}{3327}           &\multicolumn{2}{c}{19717}          &19793             &18333          &34493          &13381          &7487\\
    \multicolumn{2}{c}{\textbf{\#Edges}}        &\multicolumn{2}{c}{5278}           &\multicolumn{2}{c}{4552}           &\multicolumn{2}{c}{44324}          &130622            &81894          &24762          &245778         &119043\\
    \multicolumn{2}{c}{\textbf{\#Features}}     &\multicolumn{2}{c}{1433}           &\multicolumn{2}{c}{3703}           &\multicolumn{2}{c}{500}            &8710              &6805           &8415           &767            &745\\
    \multicolumn{2}{c}{\textbf{\#Density}}      &\multicolumn{2}{c}{0.0014}         &\multicolumn{2}{c}{0.0008}         &\multicolumn{2}{c}{0.0002}         &0.0003             &0.0005         &0.0004         &0.0027         &0.0042\\
    \multicolumn{2}{c}{\textbf{\#Classes}}      &\multicolumn{2}{c}{7}              &\multicolumn{2}{c}{6}              &\multicolumn{2}{c}{3}              &70                 &15             &5              &10             &8\\
    \multicolumn{2}{c}{\textbf{\#Data Split}}   &\multicolumn{2}{c}{140/500/1000}   &\multicolumn{2}{c}{120/500/1000}   &\multicolumn{2}{c}{60/500/1000}    &1400/2100/rest     &300/450/rest   &100/150/rest   &200/300/rest   &160/240/rest\\
\cmidrule(l{0pt}r{5pt}){1-2}\cmidrule(l{5pt}r{5pt}){3-4}\cmidrule(l{5pt}r{5pt}){5-6}\cmidrule(l{5pt}r{5pt}){7-8} \cmidrule(l{5pt}r{5pt}){9-9} \cmidrule(l{5pt}r{5pt}){10-10} \cmidrule(l{5pt}r{5pt}){11-11} \cmidrule(l{5pt}r{5pt}){12-12} \cmidrule(l{5pt}r{5pt}){13-13}
\textbf{Metric}&\textbf{Feature}&\textbf{Public}&\textbf{Random}&\textbf{Public}&\textbf{Random}& \textbf{Public}&\textbf{Random}&\textbf{Random}&\textbf{Random}&\textbf{Random}&\textbf{Random}&\textbf{Random}\\ \midrule
    Feature            &$\mathbf{X}$                &\ms{58.90}{1.35}	              &\ms{60.19}{0.00}	             &\ms{58.69}{1.28}	                &\ms{61.70}{0.00}	                &\ms{69.96}{2.89}	                &\ms{73.90}{0.00}             &\ms{40.06}{1.07}	&\ms{88.14}{0.26}	&\ms{87.49}{1.16}	&\ms{67.48}{1.48}	&\ms{59.52}{3.60}\\
    PCA                 &$\mathbf{X}$               &\ms{57.91}{1.36}	              &\ms{59.90}{0.00}	             &\ms{58.31}{1.46}	                &\ms{60.00}{0.00}	                &\ms{69.74}{2.79}	                &\ms{74.00}{0.00}             &\ms{38.46}{1.13}	&\ms{88.59}{0.29}	&\ms{87.66}{1.05}	&\ms{72.65}{1.43}	&\ms{57.45}{4.38}\\
    SVD                 &$\mathbf{X}$               &\ms{58.57}{1.30}	              &\ms{60.21}{0.19}	             &\ms{58.10}{1.14}	                &\ms{60.80}{0.26}	                &\ms{69.89}{2.66}	                &\ms{73.79}{0.29}             &\ms{38.64}{1.11}	&\ms{88.55}{0.31}	&\ms{87.98}{1.10}	&\ms{68.17}{1.39}	&\ms{60.98}{3.58}\\
    isomap              &$\mathbf{X}$               &\ms{40.19}{1.24}	              &\ms{44.60}{0.00}	             &\ms{18.20}{2.49}	                &\ms{18.90}{0.00}	                &\ms{62.41}{3.65}	                &\ms{63.90}{0.00}             &\ms{4.21}{0.25}	&\ms{73.68}{1.25}	&\ms{82.84}{0.81}	&\ms{72.66}{1.38}	&\ms{44.00}{6.43}\\
    LLE                 &$\mathbf{X}$               &\ms{29.34}{1.24}	              &\ms{36.70}{0.00}	             &\ms{18.26}{1.60}	                &\ms{21.80}{0.00}	                &\ms{52.82}{2.08}	                &\ms{54.00}{0.00}             &\ms{5.70}{0.38}	&\ms{72.23}{1.57}	&\ms{81.35}{1.59}	&\ms{45.29}{1.31}	&\ms{35.37}{1.82}\\
    \midrule
    DeepWalk            &$\mathbf{A}$               &\ms{74.03}{1.99}	              &\ms{73.76}{0.26}	             &\ms{48.04}{2.59}	                &\ms{51.80}{0.62}	                &\ms{68.72}{1.43}	                &\ms{71.28}{1.07}             &\ms{51.65}{0.83}	&\ms{83.25}{0.54}	&\ms{88.08}{1.45}	&\ms{86.47}{1.55}	&{\ums{76.58}{1.09}}\\
    Node2vec            &$\mathbf{A}$               &\ms{73.64}{1.94}	              &\ms{72.54}{1.12}	             &\ms{46.95}{1.24}	                &\ms{49.37}{1.53}	                &\ms{70.17}{1.39}	                &\ms{68.70}{0.96}             &\ms{50.35}{0.74}	&\ms{82.12}{1.09}	&\ms{86.77}{0.83}	&\ms{85.15}{1.32}	&\ms{75.67}{1.98}\\
    \midrule
    DeepWalk+F      &$\mathbf{X},\mathbf{A}$        &\ms{77.36}{0.97}	              &\ms{77.62}{0.27}	             &\ms{64.30}{1.01}	                &\ms{66.96}{0.30}	                &\ms{69.65}{1.84}	                &\ms{71.84}{1.15}             &{\ums{54.63}{0.74}}	&\ms{83.34}{0.53}	&\ms{88.15}{1.45}	&\ubms{ {86.49}}{1.55}	&\ms{65.97}{3.68}\\               
    Node2vec+F      &$\mathbf{X},\mathbf{A}$        &\ms{75.44}{1.80}	              &\ms{76.84}{0.25}	             &\ms{63.22}{1.50}	                &\ms{66.75}{0.74}	                &\ms{70.6}{1.36}	                &\ms{69.12}{0.96}             &\ms{54.00}{0.17}	&\ms{82.20}{1.09}	&\ms{86.86}{0.80}	&\ms{85.15}{1.33}	&\ms{65.01}{2.91}\\
    GAE                 &$\mathbf{X},\mathbf{A}$    &\ms{73.68}{1.08}	              &\ms{74.30}{1.42}	             &\ms{58.21}{1.26}	                &\ms{59.69}{3.29}	                &\ms{76.16}{1.81}	                &{\ums{80.08}{0.70}} &\ms{42.54}{2.69}	&\ms{88.88}{0.83}	&\ms{91.01}{0.84}	&\ms{37.72}{9.01}	&\ms{48.72}{5.28}\\
    VGAE                &$\mathbf{X},\mathbf{A}$    &\ms{77.44}{2.20}	              &\ms{76.42}{1.26}	             &\ms{59.53}{1.06}	                &\ms{60.37}{1.40}	                &\ms{78.00}{1.94}	                &\ms{77.75}{0.77}             &\ms{53.69}{1.32}	&\ms{88.66}{1.04}	&\ms{90.33}{1.77}	&\ms{49.09}{5.95}	&\ms{48.33}{1.74}\\
    DGI                 &$\mathbf{X},\mathbf{A}$    &\ms{81.26}{1.24}	              &\ms{82.11}{0.25}	             &{\ums{69.50}{1.29}}	    &\ms{70.15}{1.10}	                &\ms{77.70}{3.17}	                &\ms{79.06}{0.51}             &\ms{53.89}{1.38}	&{\ums{91.22}{0.48}}	&{\ums{92.12}{1.29}}	&\ms{79.62}{3.31}	&\ms{70.65}{1.72}\\
    GRACE               &$\mathbf{X},\mathbf{A}$    &\ms{80.46}{0.05}	              &\ms{80.36}{0.51}                &\ms{68.72}{0.04}	                &\ms{68.04}{1.06}	                &{\ums{80.67}{0.04}}	    &OOM                        &\ms{53.95}{0.11}	&\ms{90.04}{0.11}	&OOM        	&\ms{81.94}{0.48}	&\ms{70.38}{0.46}\\
    GraphCL             &$\mathbf{X},\mathbf{A}$    &{\ums{81.89}{1.34}}	  &\ms{81.12}{0.04}	             &\ms{68.40}{1.07}	                &\ms{69.67}{0.13}	                &OOM        	                &\ms{81.41}{0.10}             &OOM        	&OOM        	&OOM        &\ms{79.90}{2.05}	&OOM\\
    GMI                 &$\mathbf{X},\mathbf{A}$    &\ms{80.28}{1.06}	              &{\ums{81.20}{0.78}}    &\ms{65.99}{2.75}	                &{\ums{70.50}{0.36}}     &OOM        	                &OOM                        &OOM        	&OOM        	&OOM        	&\ms{52.36}{5.22} &OOM\\
    \midrule
    \textbf{$\mathrm{G}^2\mathrm{R}$(ours)} &$\mathbf{X},\mathbf{A}$    &{\ubms{82.58}{1.41}}	  &{\ubms{83.32}{0.75}}       &{\ubms{71.2}{1.01}}	        &{\ubms{70.66}{0.49}}	    &{\ubms{81.69}{0.98}}	    &{\ubms{81.69}{0.42}}    &{\ubms{59.70}{0.59}}	    &{\ubms{92.64}{0.40}}	&{\ubms{94.93}{0.07}} &\ms{82.24}{0.71}	&{\ubms{90.68}{0.31}}	\\ 
    \bottomrule
    \end{tabular}
    \vspace{-8pt}
\end{table*}

\section{Experiments}
In this section, we conduct experiments with synthetic graph and real-world graphs to comprehensively evaluate $\mathrm{G}^2\mathrm{R}$. The main observations in experiments are highlighted as \Circled{\footnotesize \#} \textbf{boldface}.

\subsection{What is \texorpdfstring{$\mathrm{G}^2\mathrm{R}$}{} Doing? Empirical Verification with Synthetic Graph Data}\label{sec:exp:syn}
We experiment with a synthetic graph to empirically verify that $\mathrm{G}^2\mathrm{R}$ tends to project node representations in different communities into different subspaces. The results are presented in Figure~\ref{fig:synthetic}.

\vspace{-5pt}\subsubsection{Synthetic Graph Generation.}
\begin{figure}[!t]
    \includegraphics[width=0.48\textwidth]{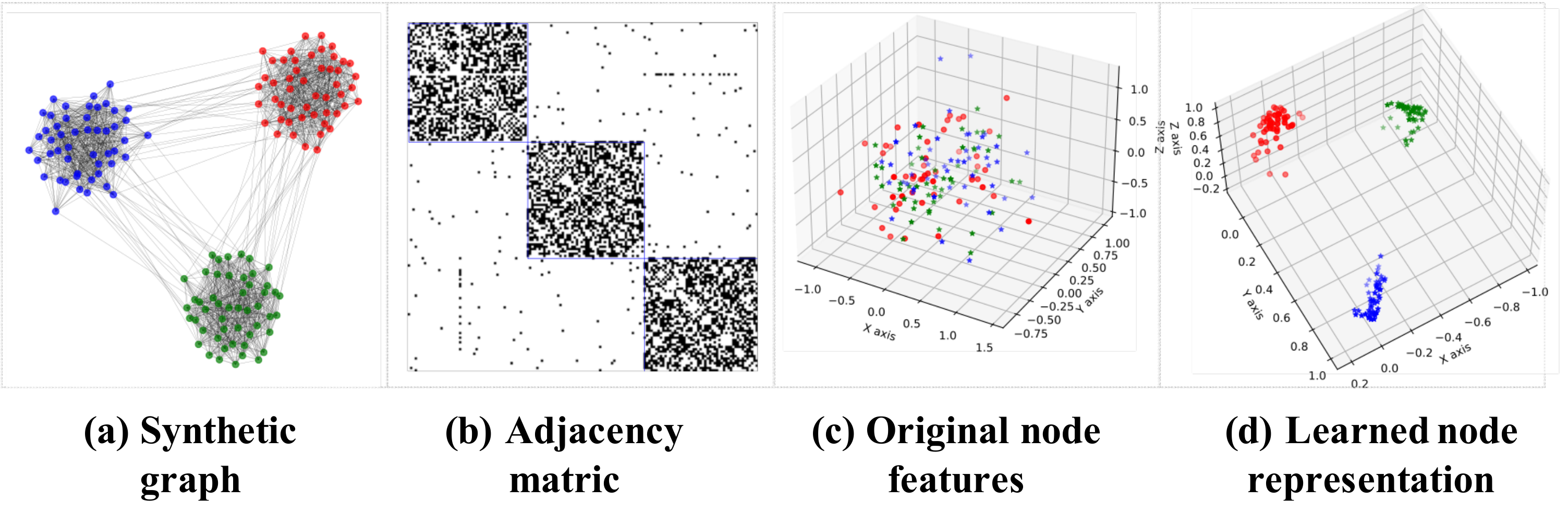}
    \vspace{-20pt}
    \caption{ Synthetic graph and visualization of its node features and representations. The different colors in (a)(c)(d) indicate different communities. The learned node representations in (d) are 3-dimensional vectors obtained by $\mathrm{G}^2\mathrm{R}$.}\label{fig:synthetic}
  \vspace{-8pt}
\end{figure}

The synthetic graph is generated as follows: i) Graph structure. We partition nodes into $3$ balanced communities and construct edges with Gaussian random partition\footnote{https://networkx.org}. The nodes within the same community have a high probability $0.5$ to form edges and a lower probability $0.01$ for nodes in different communities.
Figure~\ref{fig:synthetic}(a) and Figure~\ref{fig:synthetic}(b) show the structure of the synthetic graph and its adjacency matrix, respectively.
ii) Node features. The node feature is generated from multivariate Gaussian distributions with the same mean and standard deviation, the dimension of which is $32$. t-SNE~\cite{van2008visualizing} of node features to 3-dimensional space are in Figure~\ref{fig:synthetic}(c). Figure~\ref{fig:synthetic}(d) is the visualization of the learned node representations, the dimension of which is $3$.

\vspace{-5pt}\subsubsection{Results}
Comparing Figures~\ref{fig:synthetic}(c) and~\ref{fig:synthetic}(d), we observed that \Circled{\footnotesize 1}~\textbf{the learned node representations in different communities are nearly orthogonal} in the three-dimensional space. Moreover, we also compute the cosine similarity between each pair of the node representations to quantify the geometric relation and we observe that the cosine similarity scores for node representations pair between the different communities are extremely close to $0$. 
This observation indicates that $\mathrm{G}^2\mathrm{R}$ tends to maximize the principal angle of representation spaces of nodes in different communities. \Circled{\footnotesize 2}~\textbf{The node representations in the same community are compact.} Figure~\ref{fig:synthetic}(c) shows the original features of nodes in the same color are loose while node representations in Figure~\ref{fig:synthetic}(d) in the same color cluster together. This observation shows that $\mathrm{G}^2\mathrm{R}$ can compact the node representations in the same community. The experimental results on synthetic data are remarkably consistent with the theoretical analysis in Section~\ref{sec:theo:doing} that the node representations in different communities will be (nearly) orthogonal.

\subsection{Will \texorpdfstring{$\mathrm{G}^2\mathrm{R}$}{}{} Perform Better than Unsupervised Counterparts?}\label{sec:exp:unsuper}
We contrast the performance of the node classification task of $\mathrm{G}^2\mathrm{R}$ and various unsupervised baselines. 

\vspace{-5pt}\subsubsection{Experiment Setting}
\emph{For dataset}, we experiment on eight real-world datasets, including citation network~\cite{yang2016revisiting, bojchevski2018deep}~(Cora, CiteSeer, PubMed, CoraFull), co-authorship networks~\cite{shchur2018pitfalls} (Physics, CS), and Amazon co-purchase networks~\cite{mcauley2015image} (Photo, Computers). The details of datasets are provided in Appendix~\ref{sec:app:unsuperbaseline}.
\emph{For baselines}, we compare three categories of unsupervised baselines. The first category only utilizes node features, including original node features, PCA~\cite{wold1987principal}, SVD~\cite{golub1971singular}, LLE~\cite{roweis2000nonlinear} and Isomap~~\cite{tenenbaum2000global}. The second only considers adjacency information, including DeepWalk~\cite{perozzi2014deepwalk} and Node2vec~\cite{grover2016node2vec}. The third considers both, including DGI~\cite{velivckovic2018deep}, GMI~\cite{peng2020graph}, GRACE~\cite{you2020graph} and GraphCL~\cite{you2020graph}. 
\emph{For evaluation}, we follow the linear evaluation scheme adopted by~\cite{velivckovic2018deep, zhu2020deep}, which first trains models in an unsupervised fashion and then output the node representations to be evaluated by a logistic regression classifier~\cite{sklearn_api}. We use the same random train/validation/test split as~\cite{fey2019fast,liu2020towards}. To ensure a fair comparison, we use 1)the same logistic regression classifier, and 2)the same data split for all models. The results are summarized in Table~\ref{tab:unsupervised:random}.

\vspace{-5pt}\subsubsection{Results} From Table~\ref{tab:unsupervised:random}, we observe that \Circled{\footnotesize 3}~\textbf{$\mathrm{G}^2\mathrm{R}$ outperforms all baselines by significant margins on seven datasets among eight dataset.} Except for the Photo dataset, $\mathrm{G}^2\mathrm{R}$ achieves the state-of-the-art performance by significant margins. The average percentage of improvement to DGI (representative unsupervised method) and GRACE (representative contrastive learning method) is 5.1\% and 5.9\%, respectively. 
Moreover, $\mathrm{G}^2\mathrm{R}$ is capable of handling large graph data. The reason is partial leverage of adjacency matrix in each training batch requires lower memory usage and less time.

\subsection{Will Representation learned by \texorpdfstring{$\mathrm{G}^2\mathrm{R}$}{} (nearly) orthogonal? Visualization Analysis}\label{sec:exp:vis}
We perform a visualization experiment to analyze the representations learned by $\mathrm{G}^2\mathrm{R}$ to verify its effectiveness further. The visualization of nodes representations of different classes is in Figure~\ref{fig:orth}.

\vspace{-5pt}\subsubsection{Results} 
\begin{figure}[!t]
     \centering
     \includegraphics[width=0.48\textwidth]{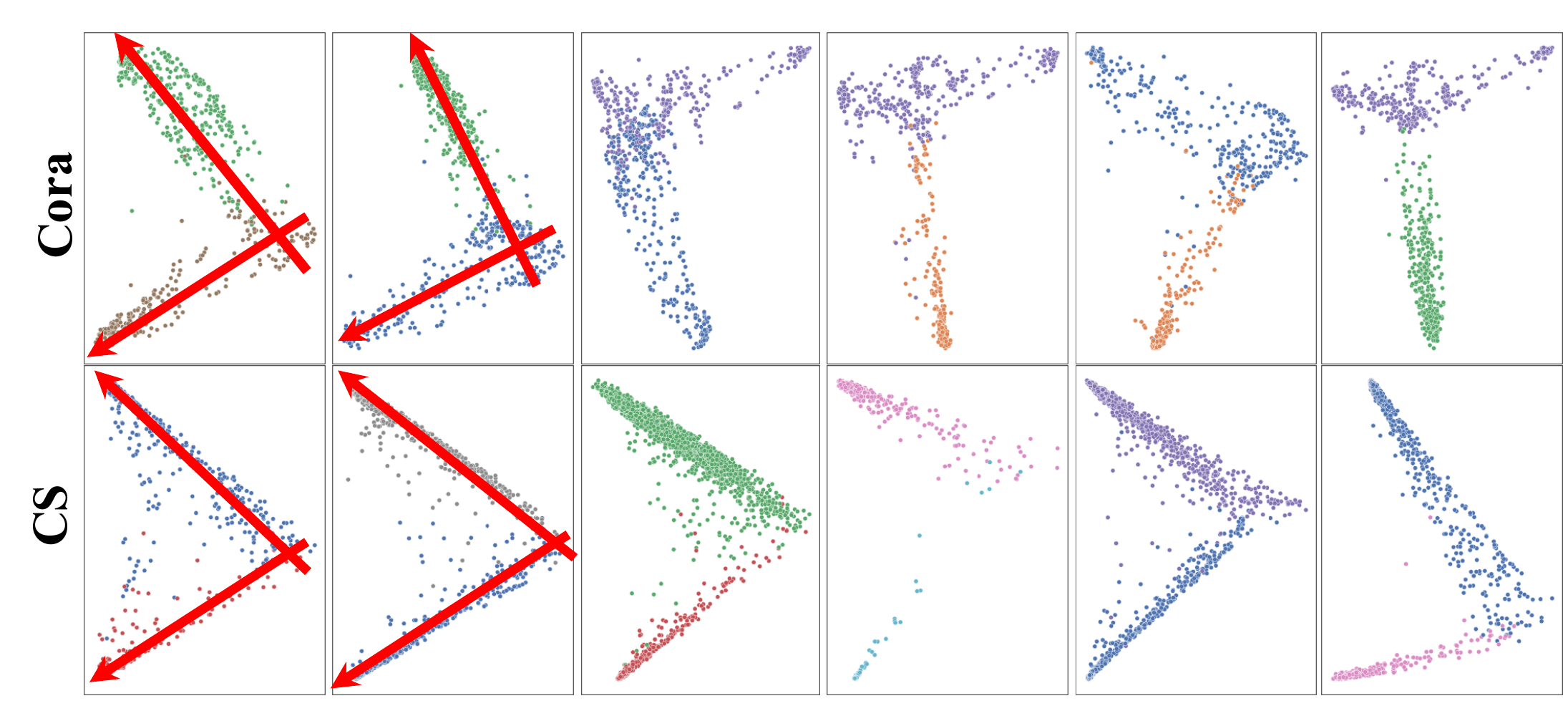}
     \vspace{-20pt}
    \caption[Caption for LOF]{PCA\footnotemark~visualization of node representations learned by $\mathrm{G}^2\mathrm{R}$ on Cora and CS dataset. Every figure only has two classes of nodes and they are nearly orthogonal. Different colors represent different classes. The direction of $\reaarr$ in the first two columns show the (nearly) orthogonality of node representations in the two classes.  }\label{fig:orth}
  \vspace{-10pt}
\end{figure}

Figure~\ref{fig:orth} remarkably shows that \Circled{\footnotesize 4}~\textbf{the representations of nodes in different classes learned by $\mathrm{G}^2\mathrm{R}$ are nearly orthogonal to each other.} Since the nodes in the same class typically connected densely, leading them to be nearly orthogonal to each other according to the proof in Section~\ref{sec:theo}. This observation also strongly supports our theoretical analysis Section~\ref{sec:theo}.

\vspace{-5pt}

\subsection{What is the Effect of Encoders and Objective Functions? Ablation Studies}\label{sec:exp:Ablation Study}
\begin{figure}[!t]
    \centering
    \includegraphics[width=0.48\textwidth]{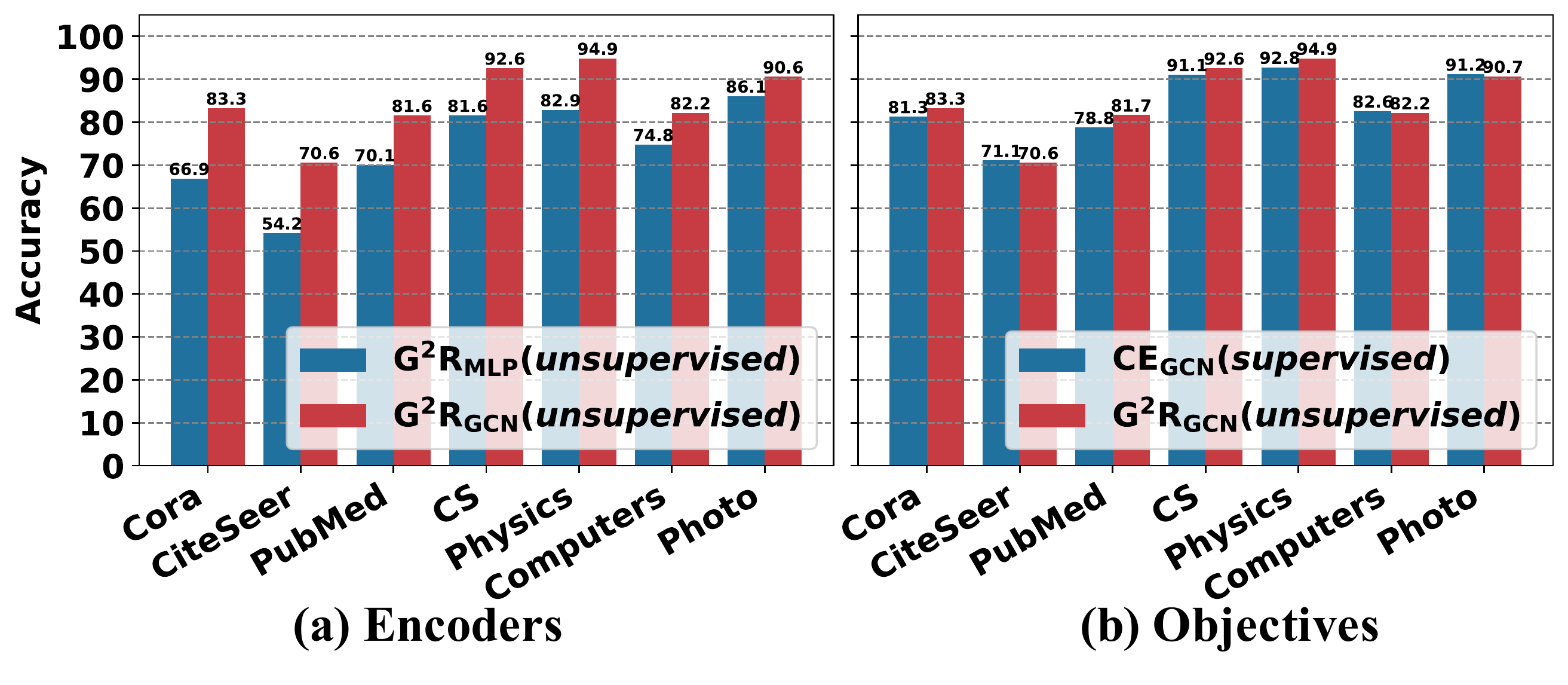}
    \vspace{-20pt}
    \caption{Ablation studies. \underline{(a)}: We compare the performance of different encoders by instantiating the encoder as GCN or MLP, respectively. \underline{(b)}: we compare the performance of different objective functions, Cross-Entropy(CE, supervised method for node classification) and ours (unsupervised).}\label{fig:as}
    \vspace{-8pt}
\end{figure}

We investigate the effect of encoder and objective function in $\mathrm{G}^2\mathrm{R}$ using ablation studies. Specifically, we replace the graph neural networks in $\mathrm{G}^2\mathrm{R}$ with other encoders or replace the proposed objective functions with cross-entropy. The results are in Figure~\ref{fig:as}.

\footnotetext{The reason why we choice PCA here is that PCA will preserve the orthogonality between vectors when transform the high-dimensional vectors to low-dimension ~\cite{jolliffe1995rotation}. Each figure includes two classes of node since we display node representation in two-dimensional space.}

\vspace{-5pt}\subsubsection{Results} 
Figures~\ref{fig:as}(a) and ~\ref{fig:as}(b) show that \Circled{\footnotesize 5} \textbf{the superiority of $\text{$\mathrm{G}^2\mathrm{R}$}_{\text{GCN}}$ is attributed to the graph neural network and the proposed objective.} Figure~\ref{fig:as}(a) indicates that graph neural networks as the encoder significantly improve the effectiveness of $\mathrm{G}^2\mathrm{R}$. Figure~\ref{fig:as}(b) shows that performance of $\mathrm{CE}_{\text{GCN}}$ drops significantly compared to the $\text{$\mathrm{G}^2\mathrm{R}$}_{\text{GCN}}$ even though the it is a supervised method for node classification. This observation indicates that superoity of $\text{$\mathrm{G}^2\mathrm{R}$}$ largely stems from the proposed objective function.

\subsection{Will the Graph Structure be Preserved in the Learned Representation?}
To investigate whether the learned node representations preserves the graph structure, we perform two visualization experiments, including \textbf{\underline{1)}} t-SNE~\cite{van2008visualizing} visualization of the original features and the node representations learned by different methods in Figure~\ref{fig:tsne}, and \textbf{\underline{2)}} visualization of the adjacency metrics of graphs and cosine similarity between learned node representations $\mathbf{Z}$ in Figure~\ref{fig:adj:vis}.
\vspace{-5pt}\subsubsection{Results} 
\begin{figure}[!t]
     \centering
     \includegraphics[width=0.48\textwidth]{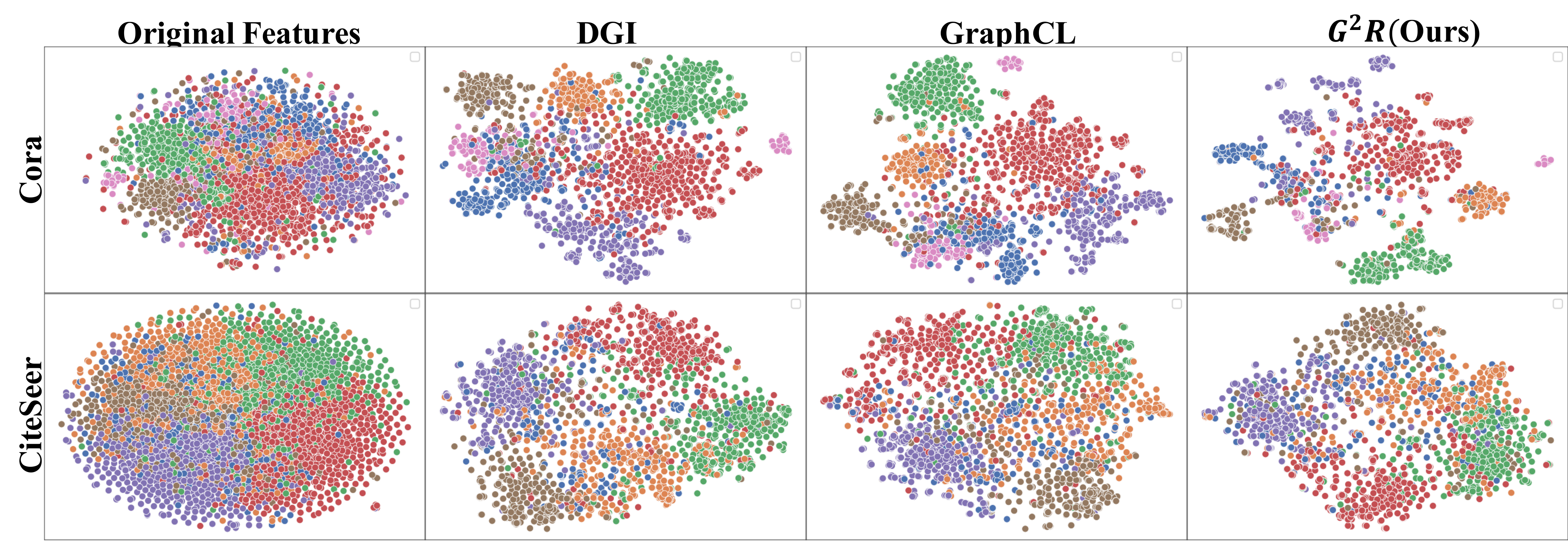}
     \vspace{-20pt}
     \caption{t-SNE~\cite{van2008visualizing} visualization of original node features and node representations learned by different methods. 
     Different colors represent different classes.
     }\label{fig:tsne}
  \vspace{-10pt}
\end{figure}

\begin{figure}[!t]
    \centering
    \includegraphics[width=0.48\textwidth]{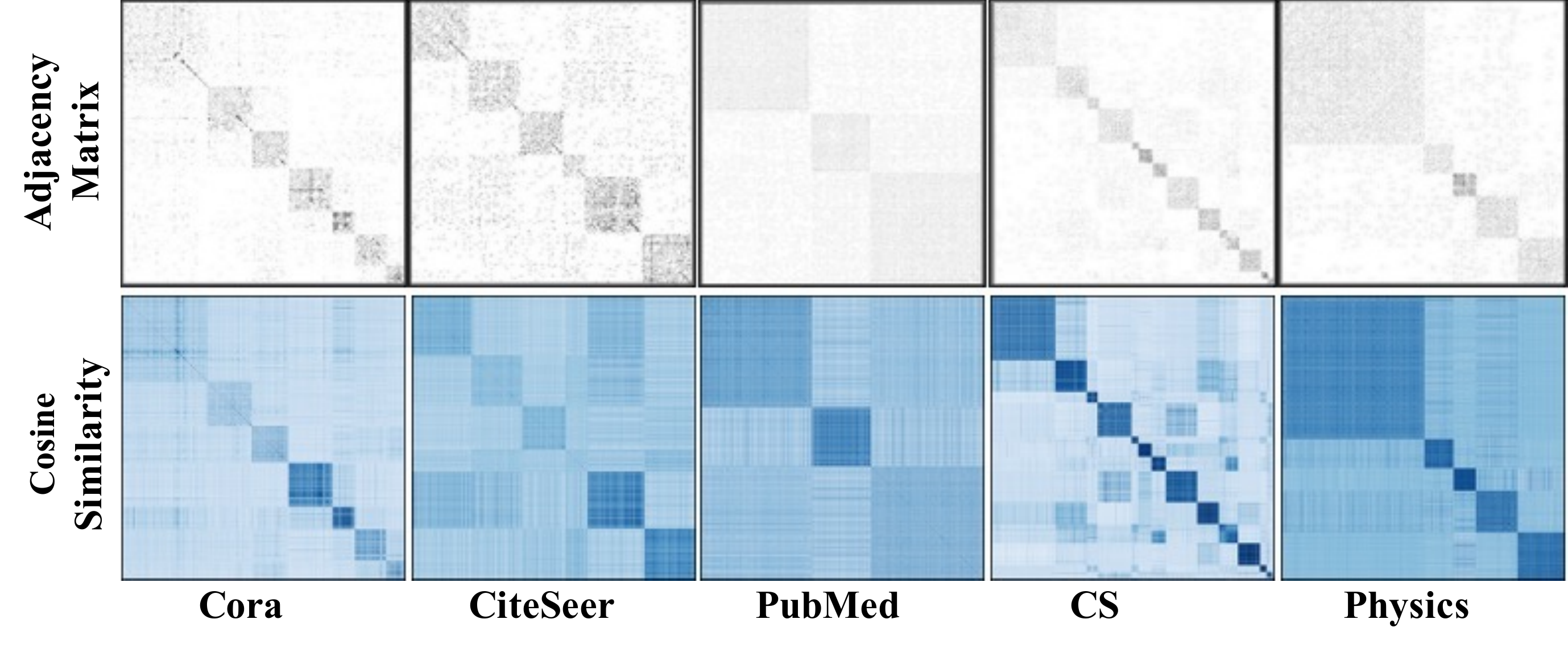}
    \vspace{-20pt}
    \caption{Visualization of adjacency matrices and the cosine similarity of learned node representations. Notably, cosine similarity can visually "recover" the adjacency matrix.
    }\label{fig:adj:vis}
  \vspace{-7pt}
\end{figure}

From Figure~\ref{fig:tsne}, \Circled{\footnotesize 6}\textbf{ the distinctly separable clusters demonstrate the discriminative capability of $\mathrm{G}^2\mathrm{R}$}. The node representations learned by $\mathrm{G}^2\mathrm{R}$ are more compact within class, leading to the discriminative node representations. The reason is that $\mathrm{G}^2\mathrm{R}$ can map the nodes in different communities into different subspaces and maximize the difference of these subspaces.
Figure~\ref{fig:adj:vis} shows that \Circled{\footnotesize 7}\textbf{ $\mathrm{G}^2\mathrm{R}$ is able to map the nodes representations in the different communities to different subspace and thus implicitly preserve the graph structure.} 
The cosine similarity of the node representations can noticeably "recover" the adjacency matrix of the graph, demonstrating the learned node representations preserved the graph structure.
\vspace{-5pt}

\vspace{-5pt}

\subsection{Will Learned Representation Perform Well on Community Detection? A Case Study}
We conduct community detection on the Cora dataset using the learned node representations by $\mathrm{G}^2\mathrm{R}$. 
\vspace{-5pt}\subsubsection{Experimental Setting}
We conduct community detection by applying K-Means to node representations learned by $\mathrm{G}^2\mathrm{R}$ and use the predicted cluster labels as communities. We use traditional community detection methods as baselines, including asynchronous fluid communities algorithm \cite{pares2017fluid} and spectral clustering \cite{ng2002spectral}. We also use the node representations learned by other unsupervised methods as baselines. The metrics to evaluate the community detection are modularity~\cite{clauset2004finding}, coverage, performance.\footnote{The detail about these metric are presented in Appendix~\ref{sec:cdm} } 
The results are in Figure~\ref{fig:cd:line}. We also show a case of community detection in Figure~\ref{fig:cd:vis}.
\vspace{-5pt}\subsubsection{Results}
\begin{figure}[!t]
  \centering
  \includegraphics[width=0.48\textwidth]{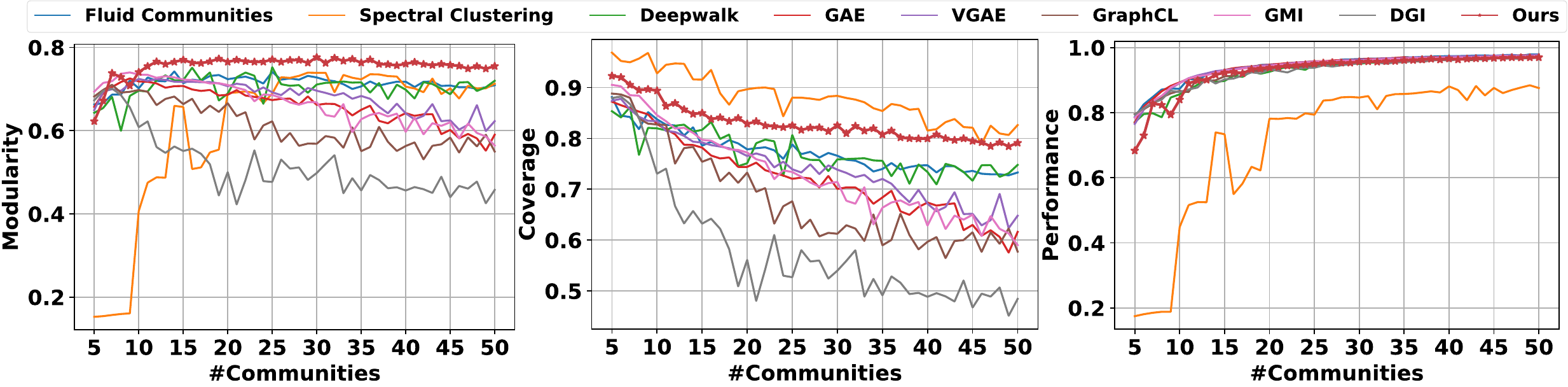}
  \vspace{-20pt}
  \caption{Performance of community detection on Cora dataset. The x-axis indicates the number of communities to be detected, the y-axis indicates the different metrics of community detection,including modularity, performance and coverage (higher is better for all metrics).}\label{fig:cd:line}
  \vspace{-10pt}
\end{figure}

\begin{figure}[!t]
  \centering
  \includegraphics[width=0.48\textwidth]{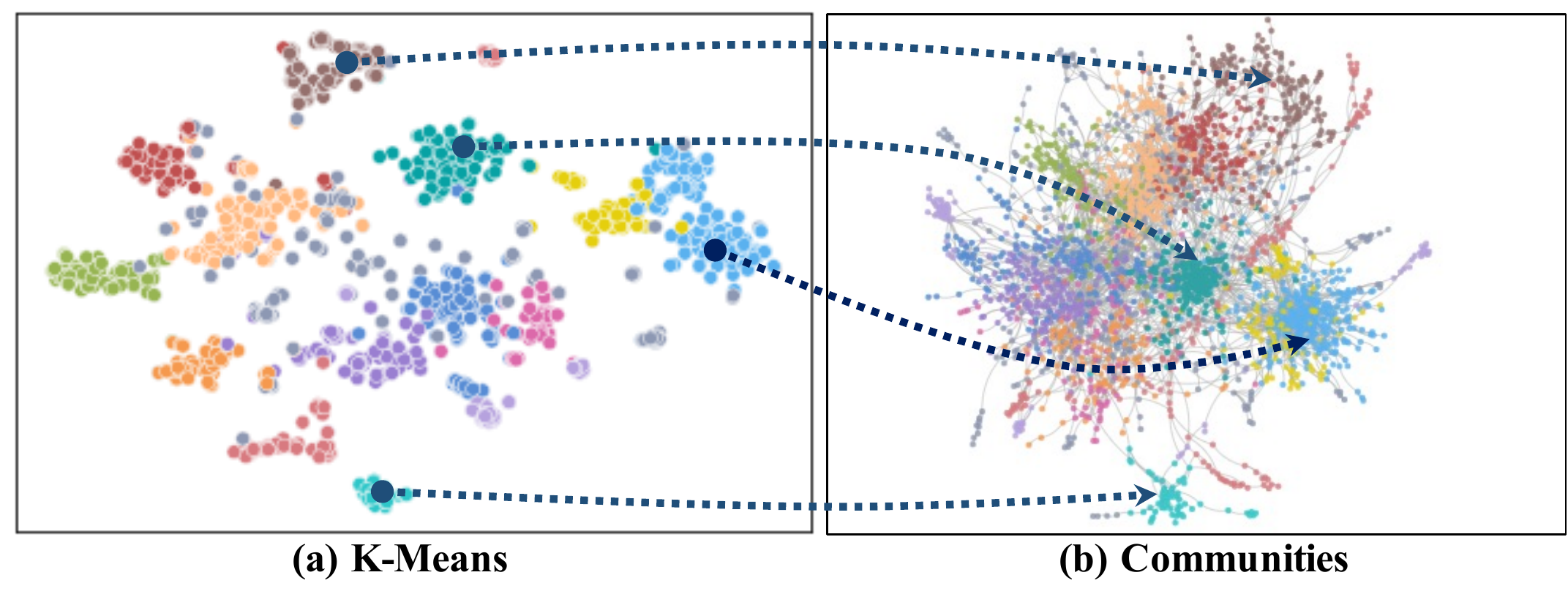}
  \vspace{-20pt}
  \caption{Community detection on Cora. We detect the communities only utilizing the learned node representations. Specifically, we use K-Means to cluster the node representations and the nodes with the same cluster label is in identical community. According to the communities label obtained from the node representations, we draw the original graph  and color the communities in the same colors to Figure~\ref{fig:cd:vis}. The same color indicates the same community.  }\label{fig:cd:vis}
  \vspace{-10pt}
\end{figure}

Figures~\ref{fig:cd:line} and~\ref{fig:cd:vis}, quantitatively and qualitatively, show \Circled{\footnotesize 8} \textbf{$\mathrm{G}^2\mathrm{R}$ outperforms the traditional community detection methods as well as unsupervised baselines for community detection task.} Figure~\ref{fig:cd:line} shows that $\mathrm{G}^2\mathrm{R}$ outperforms various community detection methods by a large margin on three metrics. In Figure~\ref{fig:cd:vis}, communities detected in Cora are visually consistent with the node representations clusters. The better performance of $\mathrm{G}^2\mathrm{R}$ results from the orthogonality of different subspaces, into which the nodes in different communities are projected.

\subsection{What is the Effect of the Hyperparameters \texorpdfstring{$\gamma_1$} and \texorpdfstring{$\gamma_2$}{}?}\label{exp:hp}
\begin{figure}[ht]
      \centering
      \includegraphics[width=0.48\textwidth]{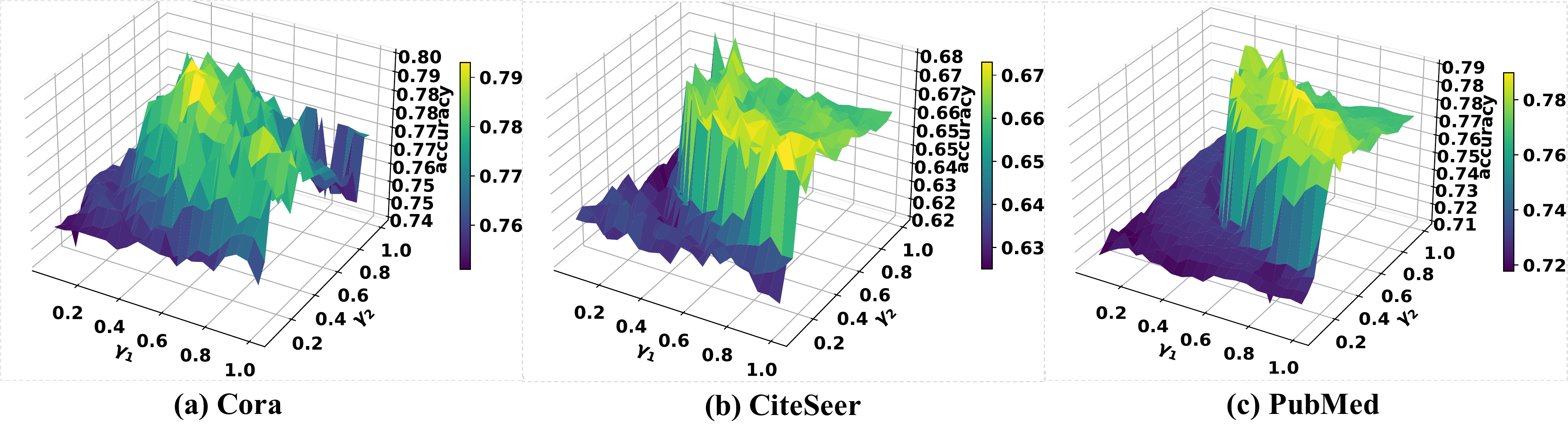}
      \vspace{-23pt}
      \caption{The effect of the hyperparameters $\gamma_1$ and $\gamma_2$ on node classification task. The x-axis indicates the values of $\gamma_1$, the y-axis indicates the values of $\gamma_2$ and the z-axis indicates the model performance on node classification task. }\label{fig:hp}
      \vspace{-15pt}
\end{figure}

We investigate the effect of hyperparameters $\gamma_1$ and $\gamma_2$ on $\mathrm{G}^2\mathrm{R}$ via training with $20$ evenly spaced values of both $\gamma_1$ and $\gamma_2$ within $(0,1]$ on Cora, CiteSeer, PubMed datasets. The results are presented in Figure~\ref{fig:hp}. From Figure~\ref{fig:hp}, we observed that \Circled{\footnotesize 9}\textbf{ hyperparameters strongly influence the performance of $\mathrm{G}^2\mathrm{R}$ and the best performance is achieved around  $\gamma_1 = \gamma_2 = 0.5$} 
The performance is lower while $\gamma_1 < 0.5$ and $\gamma_2 < 0.5$, which shows that it is important to control the dynamics of the expansion and compression of the node representations. \Circled{\footnotesize 10}\textbf{ $\mathrm{G}^2\mathrm{R}$ is not sensitive to hyperparameter across different datasets}, since $\mathrm{G}^2\mathrm{R}$ achieves the best performance with the similar hyperparameters ($\gamma_1=\gamma_2=0.5$) on Cora, CiteSeer, PubMed datasets. Based on this observation, we set $\gamma_1=\gamma_2=0.5$ on all datasets in our performance experiments.

\subsection{\texorpdfstring{$\mathrm{G}^2\mathrm{R}$}{} is even Better than Supervised Counterparts}\label{sec:exp:super}
Despite that $\mathrm{G}^2\mathrm{R}$ shows its superior performance compared to the unsupervised baselines, we contrast the performance of $\mathrm{G}^2\mathrm{R}$ and supervised methods on the node classification task. 

\vspace{-5pt}\subsubsection{Experiments Settings}
We consider the following supervised learning baselines: Logistic Regression (LogReg), Multilayer Perceptron (MLP), Label Propagation~(LP)~\cite{chapelle2009semi}, Normalized Laplacian Label Propagation (LP NL)~\cite{chapelle2009semi}, Cheb-Net~\cite{defferrard2016convolutional}, Graph Convolutional Network~(GCN)~\cite{kipf2016semi}, Graph Attention Network~(GAT)~\cite{velivckovic2018graph}, Mixture Model Network (MoNet) ~\cite{monti2017geometric}, GraphSAGE(SAGE)~\cite{hamilton2017inductive}, APPNP~\cite{klicpera2018predict}, SGC~\cite{wu2019simplifying} and DAGNN~\cite{liu2020towards}. The results of the baselines are obtained from~\cite{liu2020towards, shchur2018pitfalls}, so we follow the same data split and the same datasets in the papers~\cite{liu2020towards, shchur2018pitfalls}. We follow the linear evaluation scheme for $\mathrm{G}^2\mathrm{R}$, where $\mathrm{G}^2\mathrm{R}$ was trained in an unsupervised manner and then output the node representations as input features to a logistic regression classifier~\cite{sklearn_api}. The details of baselines are provided in Appendix \ref{sec:app:superbaseline}. The results are summarized in Table~\ref{tab:supervised}.

\vspace{-5pt}\subsubsection{Results}
From Table~\ref{tab:supervised}, we observed that \Circled{\footnotesize 11} \textbf{$\mathrm{G}^2\mathrm{R}$ shows comparable performance across all seven datasets, although the baselines are all supervised methods.} From the `Avg.rank' column in Table~\ref{tab:supervised}, $\mathrm{G}^2\mathrm{R}$ ranks $3.1$ among all the methods on all datasets. $\mathrm{G}^2\mathrm{R}$ obtains a comparable performance in node classification task even though compared to supervised baselines. This observation shows the node representations learned by $\mathrm{G}^2\mathrm{R}$ preserve the information for node classification task even though compared to the end-to-end models for the same downstream task.

\begin{table}[t]
    \centering
    \fontsize{7}{8}\selectfont  
    \setlength{\tabcolsep}{3pt}
    \vspace{-7pt}
    \caption{Comparison to supervised baselines with public data split on node classification. `P' means public data split while `R' means random data split. `Phy./Com./Pho.' means Physics/Computers/Photo dataset. The `Avg.Rank' is the average rank among all the methods on all datasets. }\label{tab:supervised}
\begin{tabular}{lccccccccccc}
\toprule
\multirow{2}{*}{\textbf{Methods}}&\multicolumn{2}{c}{\textbf{Cora}}&\multicolumn{2}{c}{\textbf{CiteSeer}}&\multicolumn{2}{c}{\textbf{PubMed}}  &\multirow{2}{*}{\textbf{CS}} &\multirow{2}{*}{\textbf{Phy.}} & \multirow{2}{*}{\textbf{Com.}} &\multirow{2}{*}{\textbf{Pho.} } &\textbf{Avg.} \\ 
\cmidrule(l{5pt}r{5pt}){2-3}\cmidrule(l{5pt}r{5pt}){4-5}\cmidrule(l{5pt}r{5pt}){6-7}
&\textbf{P}&\textbf{R}&\textbf{P}&\textbf{R}& \textbf{P}&\textbf{R} &&&& &\textbf{Rank}\\ \midrule
LogReg          &52.0 &58.3 &55.8 &60.8 &73.6 &69.7 &86.4 &86.7 &64.1 &73.0  &\textbf{11.3}\\
MLP             &61.6 &59.8 &61.0 &58.8 &74.2 &70.1 &88.3 &88.9 &44.9 &69.6  &\textbf{10.9}\\
LP              &71.0 &79.0 &50.8 &65.8 &70.6 &73.3 &73.6 &86.6 &70.8 &72.6  &\textbf{11.2}\\
LP NL           &71.2 &79.7 &51.2 &66.9 &72.6 &77.8 &76.7 &86.8 &75.0 &83.9  &\textbf{9.5}\\
ChebNet         &80.5 &76.8 &69.6 &67.5 &78.1 &75.3 &89.1 &-    &15.2 &25.2  &\textbf{10.0}\\
GCN             &81.3 &79.1 &71.1 &68.2 &78.8 &77.1 &91.1 &92.8 &82.6 &91.2  &\textbf{5.7}\\
GAT             &83.1 &80.8 &70.8 &68.9 &79.1 &77.8 &90.5 &92.5 &78.0 &85.7  &\textbf{5.8}\\
MoNet           &79.0 &84.4 &70.3 &71.4 &78.9 &83.3 &90.8 &92.5 &83.5 &91.2  &\textbf{4.0}\\
SAGE            &78.0 &84.0 &70.1 &71.1 &78.8 &79.2 &91.3 &93.0 &82.4 &91.4  &\textbf{4.7}\\
APPNP           &83.3 &81.9 &71.8 &69.8 &80.1 &79.5 &90.1 &90.9 &20.6 &30.0  &\textbf{6.0}\\
SGC             &81.7 &80.4 &71.3 &68.7 &78.9 &76.8 &90.8 &-    &79.9 &90.7  &\textbf{5.9}\\
DAGNN           &84.4 &83.7 &73.3 &71.2 &80.5 &80.1 &92.8 &94.0 &84.5 &92.0  &\textbf{1.7}\\\midrule
\textbf{Ours}   &83.3 &82.6 &70.6 &71.2 &81.7 &81.7 &92.6 &94.9 &82.2 &90.7  &\textbf{3.1}\\ 
\bottomrule
\end{tabular}
\vspace{-10pt}
\end{table}

\section{Related Works}\label{sec:related}
%
\noindent\textbf{Graph representation learning with random walks.} Many approaches~\cite{grover2016node2vec,perozzi2014deepwalk,tang2015line,qiu2018network} learn the node representations based on random walk sequences. Their key innovation is optimizing the node representations so that nodes have similar representations if they tend to co-occur over the graph. In our experiment, we use DeepWalk and node2vec as baselines, which are the representative methods based on random walk. DeepWalk~\cite{perozzi2014deepwalk}, as pioneer work to learn representations of vertices in a network, uses local information from truncated random walks as input to learn a representation which encodes structural regularities. node2vec~\cite{grover2016node2vec} aims to map nodes into a low-dimensional space while maximizing the likelihood of preserving nodes neighborhoods.

\noindent\textbf{Contrastive graph representation learning.} 
Contrastive learning is the key component to word embedding methods~\cite{collobert2008unified, mikolov2013distributed}, and recently it is used to learn representations for graph-structured data~\cite{perozzi2014deepwalk, grover2016node2vec, kipf2016semi, hamilton2017inductive, garcia2017learning}. For example, DGI~\cite{velivckovic2018deep} learns node representations in an unsupervised manner by maximizing mutual information between patch representations and the graph representation. GRACE~\cite{you2020graph} maximizes the agreement of node representations in two generated views. GraphCL~\cite{you2020graph} learns representations with graph data augmentations.

\noindent\textbf{Graph Neural Networks.}
Graph neural networks have became the new state-of-the-art approach to process graph data~\cite{hamilton2017inductive,huang2019graph}. Starting with the success of GCN in the semi-supervised node classification task\cite{kipf2016semi}, a wide variety of GNN variants have proposed for graph learning task~\cite{hamilton2017inductive,velivckovic2018graph,wu2019simplifying,gao2019graph,velivckovic2018deep}. Most of them follow a message passing strategy to learn node representations over a graph. Graph Attention Network~(GAT)~\cite{velivckovic2018graph} proposes masked self-attentional layers that allow weighing nodes in the neighborhood differently during the aggregation step. GraphSAGE~\cite{hamilton2017inductive} focuses on inductive node classification with different neighbor sampling strategies. Simple Graph Convolution~(SGC)~\cite{wu2019simplifying} reduces the excess complexity of GCNs by removing the nonlinearities between GCN layers and collapsing the resulting function into a single linear transformation. Personalized propagation of neural predictions (PPNP) and (APPNP)~\cite{klicpera2018predict} leverage adjustable neighborhood for classification and can be easily combined with any neural network. However, all these methods are typically supervised, which highly rely on reliable labels. In this work, we leverage the graph neural network to encode the graph to node representations.

\section{Conclusion}\label{sec:conlcu}
Graph representation learning becomes a dominant technique in analyzing graph-structured data. 
In this work, we propose Geometric Graph Representation Learning ($\mathrm{G}^2\mathrm{R}$), an unsupervised approach to learning discriminative node representations for graphs. Specifically, we propose an objective function to enforce discriminative node representations via maximizing the principal angle of the subspace of different node groups. And we provide theoretical justification for the proposed objective function, which can guarantee the orthogonality for node in different groups. We demonstrate competitive performance of $\mathrm{G}^2\mathrm{R}$ on node classification and community detection tasks. Moreover, $\mathrm{G}^2\mathrm{R}$ even outperforms multiple supervised counterparts on node classification task. The strength of $\mathrm{G}^2\mathrm{R}$ suggests that, despite a recent surge in deeper graph neural networks, unsupervised learning on graph remains promising.

\begin{acks}
We would like to thank all the anonymous reviewers for their valuable suggestions. This work is in part supported by NSF IIS-1849085, CNS-1816497, IIS-1750074, and IIS-2006844. The views and conclusions contained in this paper are those of the authors and should not be interpreted as representing any funding agencies.
\end{acks}

\bibliographystyle{ACM-Reference-Format}
\bibliography{ref}

\appendix

\clearpage

\section{Theoretical Analysis}\label{sec:app:theo}
\subsection{Preliminaries}

\begin{theorem}\cite{miao1992principal}\label{theo:vol}
Let $\mathbf{A} = (\mathbf{A}_1, \mathbf{A}_2) $, $\mathbf{A}_1 \in \mathbb{R}_{l}^{n\times n_1 }$, $\mathbf{A}_1 \in \mathbb{R}_{m}^{n\times n_2 }$, and $\text{rank}( \mathbf{A} )=l+m$. Then
\begin{align*}
    &\text{vol}(\mathbf{A}) =\text{vol}(\mathbf{A}_1) \text{vol}(\mathbf{A_2}) \text{sin}\{ R(\mathbf{A}_1),R(\mathbf{A}_2)  \},
\end{align*}
where $\text{vol}(\mathbf{A}) = \sqrt{\text{det}(\mathbf{A}^{\top}\mathbf{A}) }$ measures the compactness of $\mathbf{A}$ and $\text{sin}\{ R(\mathbf{A}_1),R(\mathbf{A}_2)  \}$ is the product of principal sines between $R(\mathbf{A}_1)$ and $R(\mathbf{A}_2)$.
\end{theorem}

\begin{corollary}\label{corollary:mono}
Let $\mathbf{I} + a\mathbf{Z}^{\top}\mathbf{Z}$ and $\mathbf{Z} = [\mathbf{z}_1,\mathbf{z}_2,\cdots,\mathbf{z}_n]\in\mathbb{R}^{d\times n}$,  while the $\mathbf{z}_i$ in $\mathbf{Z}$ are pairwise orthogonal, then the $\Tilde{\mathbf{z}}_i$ in $\Tilde{\mathbf{Z}}$ are pairwise orthogonal.
\end{corollary}

\textbf{Proof}:
Suppose $\mathbf{Z} = \mathbf{U}\mathbf{\Sigma}\mathbf{V}^{\top}$, then we have
\begin{equation}
    \mathbf{I} + a\mathbf{Z}^{\top}\mathbf{Z} = \mathbf{I} + a\mathbf{U}\mathbf{\Sigma}^2\mathbf{V}^{\top} =\mathbf{U}(I + a\mathbf{\Sigma}^2)\mathbf{V}^{\top} =\mathbf{ \Tilde{Z} }^{\top} \mathbf{ \Tilde{Z} }.
\end{equation}
We can see from the above derivation, while the $\mathbf{z}_n$ in $\mathbf{Z}$ are pairwise orthogonal, the result of $\mathbf{Z}^{\top}\mathbf{Z}$ is a diagonal matrix, then $\mathbf{U}\mathbf{\Sigma}^2\mathbf{V}^{\top}$ is diagonal matrix, thus $\mathbf{ \Tilde{Z} }^{\top} \mathbf{ \Tilde{Z} }=\mathbf{U}(I + a\mathbf{\Sigma}^2)\mathbf{V}^{\top}$ is diagonal matrix. So the $\Tilde{\mathbf{z}}_n$ in $\Tilde{\mathbf{Z}}$ are pairwise orthogonal.

\subsection{Insights of Coding Rate.}

We first present how to derive the coding rate of entire node representations following \cite{ma2007segmentation}.

Suppose we have data $\mathbf{W}= (w_1, w_2, \cdots, w_m)$, and let $\epsilon^2$ be the error allowable for encoding every vector $w_i$ in $\mathbf{W}$. In other words, we are allowed to distort each vector of $w_i$ with random variable $z_i$ of variance $\epsilon^2/n$. So we have  
\begin{equation}
\begin{split}
    &\hat{w}_i = w_i + z_i,\text{with}~z_i = \mathcal{N}(0,\frac{\epsilon^2}{n}\mathbf{I}),
\end{split}
\end{equation}
Then the covariance matrix of $w_i$ is
\begin{equation}
\begin{split}
    &\hat{\Sigma} \doteq \mathbb{E}[\frac{1}{m} \sum_{i=1}^{m}\hat{w}_i\hat{w}_i^{\top} ] = \frac{\epsilon^2}{n}\mathbf{I}+\frac{1}{m}\mathbf{W}\mathbf{W}^{\top},
\end{split}
\end{equation}
And the volumes of covariance matrix and random vector $z_i$ are
\begin{equation}
\begin{split}
    &\text{vol}(\hat{\mathbf{W}}) \propto \sqrt{ \text{det}(\frac{\epsilon^2}{n}\mathbf{I}+\frac{1}{m}\mathbf{W}\mathbf{W}^{\top}) },\\
    &\text{vol}(z) \propto \sqrt{ \text{det}(\frac{\epsilon^2}{n}\mathbf{I}) },
\end{split}
\end{equation}
Then the number of bit needed to encode the data $\mathbf{W}$ is
\begin{equation}
\begin{split}
    &R(\mathbf{W}) = \text{log}_2( \frac{\text{vol}(\hat{\mathbf{W}})}{ \text{vol}(z) } ) = \frac{1}{2}\text{log}_2 \text{det}( \mathbf{I}+\frac{n}{m\epsilon^2}\mathbf{W}\mathbf{W}^{\top}).
\end{split}
\end{equation}

\subsection{Proof of Equation~\texorpdfstring{\eqref{equ:ssbmloss}}{} }\label{sec:app:proof}
We take  $\mathbf{I} + \frac{d}{N\epsilon^2}\mathbf{Z}^{\top}\mathbf{Z} =\mathbf{ \Tilde{Z} }^{\top} \mathbf{ \Tilde{Z} }$ and $\beta =  \text{sin}\{ R(\mathbf{\Tilde{Z}}_1),R( \mathbf{\Tilde{Z}}_{2}) \}$, then we have

\begin{scriptsize}
\begin{equation}
\begin{split}
    &\Delta R_{\mathcal{G}}( \mathbf{Z}, \mathcal{C}, \epsilon ) \notag\\
    &= R_{\mathcal{G}}( \mathbf{Z},\epsilon ) -  R^c_{\mathcal{G}}( \mathbf{Z}, \epsilon |\mathcal{C}  ) \notag\\
    &=\frac{1}{2}\text{log}\text{det} \left( \mathbf{I} + \frac{d}{N\epsilon^2}\mathbf{Z}^{\top}\mathbf{Z} \right) - \sum^{2M}_{j=1}\frac{tr(\mathbf{C}_j)}{2N}\text{log}\text{det} \left( \mathbf{I} + \frac{d}{tr(\mathbf{C}_j)\epsilon^2}\mathbf{Z}^{\top}\mathbf{C}_j\mathbf{Z} \right) \notag\\
    &=\frac{1}{2}\text{log}\text{det} \left( \mathbf{I} + \frac{d}{N\epsilon^2}\mathbf{Z}^{\top}\mathbf{Z} \right) - \frac{1}{M}\sum^{2}_{j=1}\frac{tr(\mathbf{C}^j)}{2N}\text{log}\text{det} \left( \mathbf{I} + \frac{d}{tr(\mathbf{C}^j)\epsilon^2}\mathbf{Z}^{\top}\mathbf{C}^j\mathbf{Z} \right) \notag\\
    &=\frac{1}{2}\text{log}\text{det} \left( \mathbf{I} + \frac{d}{N\epsilon^2}\mathbf{Z}^{\top}\mathbf{Z} \right) - \frac{1}{M}\sum^{2}_{j=1}\frac{(p^{i} - p^{o})M}{2N}\text{log}\text{det} \left( \mathbf{I} + \frac{d\cdot (p^{i} - p^{o}) }{M\cdot (p^{i} - p^{o})\cdot \epsilon^2}\mathbf{Z}_j^{\top}\mathbf{Z}_j \right) \notag\\
    &=\frac{1}{2}\text{log}\text{det} \left( \mathbf{I} + \frac{d}{N\epsilon^2}\mathbf{Z}^{\top}\mathbf{Z} \right) - \sum^{2}_{j=1}\frac{(p^{i} - p^{o})}{2N}\text{log} \text{det} \left( \mathbf{I} + \frac{d }{M\epsilon^2}\mathbf{Z}_j^{\top}\mathbf{Z}_j \right) \notag\\
    &=\frac{1}{2}\text{log} \text{det} \left(\mathbf{ \Tilde{Z} }^{\top}\mathbf{ \Tilde{Z} } \right) - \sum^{2}_{j=1}\frac{(p^{i} - p^{o})}{2N}\text{log}\text{det} \left( \mathbf{I} + \frac{d }{M\epsilon^2}\mathbf{Z}_j^{\top}\mathbf{Z}_j \right)\notag\\
    &=\frac{1}{2}\sum_{j=1}^{2}\frac{1}{2}\text{log} \text{det} \left(\mathbf{ \Tilde{Z} }_j^{\top}\mathbf{ \Tilde{Z} }_j \right) + \frac{1}{2}\cdot\text{log}\beta - \sum^{2M}_{j=1}\frac{(p^{i} - p^{o})}{2}\text{log}\text{det} \left( \mathbf{I} + \frac{d }{M\epsilon^2}\mathbf{Z}_j^{\top}\mathbf{Z}_j \right) \notag\\
    &=\sum_{j=1}^{2}\frac{1}{4}\text{log} \text{det} \left( \mathbf{I} + \frac{d}{N\epsilon^2}\mathbf{Z}_j^{\top}\mathbf{Z}_j \right)  - \sum^{2}_{j=1}\frac{(p^{i} - p^{o})}{2N}\cdot\text{log}\text{det} \left( \mathbf{I} + \frac{d }{M\epsilon^2}\mathbf{Z}_j^{\top}\mathbf{Z}_j \right) + \frac{1}{2}\cdot\text{log}\beta\notag\\
    &=\sum_{j=1}^{2}\text{log}\left( \frac{  \text{det}^{\frac{1}{4}} \left( \mathbf{I} + \frac{d}{N\epsilon^2}\mathbf{Z}_j^{\top}\mathbf{Z}_j \right) }{ \text{det}^{\frac{p^{i} - p^{o}}{2N}} \left( \mathbf{I} + \frac{d }{M\epsilon^2}\mathbf{Z}_j^{\top}\mathbf{Z}_j \right)} \right) + \frac{1}{2}\cdot\text{log}\beta.
\end{split}
\end{equation}
\end{scriptsize}
The $\beta = \text{sin}\{ R(\mathbf{\Tilde{Z}}_1),R( \mathbf{\Tilde{Z}}_{2}) \}$ means the principal angle of the $\mathbf{\Tilde{Z}}_1,\mathbf{\Tilde{Z}}_{2}$, which measures the difference of subspaces. Maximizing $\beta$ is to maximize the difference of the subspace. According to Corollary 1, we prove that the $\mathbf{z}_*$ in $\mathbf{Z}$ are pairwise orthogonal, then the $\Tilde{\mathbf{z}}_*$ in $\Tilde{\mathbf{Z}}$ will also be pairwise orthogonal. So the maximum value of the product of principal angle sines between different subspaces of $\mathbf{Z}$ and $\Tilde{\mathbf{Z}}$ are equal to $1$. And then they reach the maximum at the same time.

\section{Experimental Setting}\label{sec:app:exp}
To reproduce the results of the proposed method, we provide the details of training, dataset, baselines.

\subsection{Training Setting}
$\mathrm{G}^2\mathrm{R}$ is implemented using PyTorch 1.7.1~\cite{paszke2019pytorch} and PyTorch Geometric 1.6.3~\cite{fey2019fast}.
All models are initialized with Xavier~\cite{glorot2010understanding} initialization, and are trained with Adam~\cite{kingma2015adam} optimizer.
For linear evaluation mode for node classification, we use the existing implementation of logistic regression with $L_2$ regularization from Scikit-learn~\cite{sklearn_api}.
For all datasets and baselines, we perform experiments five times with different seeds and report the mean and standard deviation of accuracies (\%) for node classification. In training phase, we set the dimension of node representations as $512$. We perform grid search on the number of epoch and learning rate. For unsupervised baselines, we use the public code released by the authors. All experiments are conducted on a Linux server with two AMD EPYC 7282 CPUs and four NVIDIA RTX3090 GPUs (24GB memory each).

\subsection{Dataset}
Following the previous works, we use eight benchmark datasets to evaluate $\mathrm{G}^2\mathrm{R}$ and baselines, including Cora, CiteSeer, PubMed, CoraFull, Coauthor CS, Coauthor Physics, Amazon Computers, and Amazon Photo~\cite{yang2016revisiting,bojchevski2018deep,shchur2018pitfalls,mcauley2015image}. All datasets used throughout experiments are available in PyTorch Geometric~\cite{fey2019fast} libraries. 
The details of the dataset are as follows:

\begin{itemize}[leftmargin=0.4cm]
\item \textbf{Planetoid}~\cite{yang2016revisiting}. Planetoid dataset includes Cora, CiteSeer and PubMed, which is representative citation network datasets. These datasets contains a number of machine learning papers, where nodes and edges denote documents and citation, respectively. Node features are bay-of-words for documents. Class labels indicate the field of documents.

\item \textbf{CoraFull}~\cite{bojchevski2018deep} is a well-known citation network that contains labels based on the paper topic. This dataset is additionally extracted from the original data of the entire network of Cora. Specifically, CoraFull contains the entire citation network of Cora, while the Planetoid Cora dataset is its subset.

\item \textbf{Coauthor}~\cite{shchur2018pitfalls}. Coauthor Physics is co-authorship graph based on the Microsoft Academic Graph from the KDD Cup 2016 challenge. Nodes are authors and edges indicate co-authored a paper. Node features represent paper keywords for each author’s papers, and class labels indicate the most active fields of study for each author.

\item \textbf{Amazon}~\cite{shchur2018pitfalls}. Amazon dataset includes Computers and Photo which are extracted from co-purchase graph~\cite{mcauley2015image}. Nodes represent goods, edges indicate that two goods were bought together. The node features are bag-of-words encoded product reviews and class labels are the product category.
\end{itemize}

\subsection{Baselines for unsupervised learning}\label{sec:app:unsuperbaseline}
We list the baselines used for the unsupervised learning comparison.
\begin{itemize}[leftmargin=0.4cm]
    \item \textbf{Features}. We use the original feature as input.
    \item \textbf{PCA}~\cite{wold1987principal} and \textbf{SVD}~\cite{golub1971singular}. These two methods are matrix decomposition based methods and only contain the node features information. We use the node features after PCA( or SVD) dimensionality reduction as the input feature.
    \item \textbf{LLE}~\cite{roweis2000nonlinear} and \textbf{Isomap}~~\cite{tenenbaum2000global}. These two methods are manifold based dimensionality reduction methods and only contain the node features information. We use the reduced node feature as the node representations.
    \item \textbf{DGI}~\footnote{\url{https://github.com/rusty1s/pytorch\_geometric/blob/master/examples/infomax\_inductive.py}}~\cite{velivckovic2018deep} is a general approach for learning node representations within graph-structured data in an unsupervised manner, which relies on maximizing mutual information between patch representations and corresponding high-level summaries of graphs—both. 
    \item \textbf{GraphCL}~\footnote{\url{https://github.com/Shen-Lab/GraphCL}}~\cite{you2020graph} is a graph contrastive learning framework for learning unsupervised representations of graph data with graph data augmentations.
    \item \textbf{GRACE}~\footnote{\url{https://github.com/CRIPAC-DIG/GRACE}}~\cite{you2020graph} is an unsupervised  graph representation learning method. GRACE first generates two views of graph by corruption and then maximizes the agreement of node representations in these two views.
    \item \textbf{GMI}~\footnote{\url{https://github.com/zpeng27/GMI}}~\cite{peng2020graph} measures the correlation between input graphs and high-level hidden representations. GMI directly maximizes the mutual information between the input and output of a graph encoder in terms of node features and topological structure.
\end{itemize}

\subsection{Baselines for supervised learning}\label{sec:app:superbaseline}

We mainly adopt supervised GNN models as baselines for supervised learning comparison.  In addition to GNN models, we also consider the following baselines: Logistic Regression (LogReg), Multi-Layer Perceptron~(MLP), Label Propagation~(LabelProp) and Normalized Laplacian Label Propagation~(LabelProp NL). Then details of baseline models are listed as follows:

\begin{itemize}[leftmargin=0.4cm]

\item \textbf{MLP} uses the node features as input and the node labels as output, which only leverages the node feature information while ignores the connection information.

\item \textbf{LabelProp}~\cite{zhu2002learning} uses unlabeled data to help labeled data in classification. Labels were propagated with a combination of random walk and clamping. LabelProp only considers the graph structure.

\item \textbf{GCN}~\cite{kipf2016semi} Graph Convolutional Network is one of the earlier models that works by performing a linear approximation to spectral graph convolutions.

\item \textbf{MoNet}~\cite{monti2017geometric} generalizes the GCN architecture and allows to learn adaptive convolution filters. 

\item \textbf{GAT}~\cite{velivckovic2018graph} proposes masked self-attentional layers that allow weighing nodes in the neighborhood differently during the aggregation step, which overcomes the shortcomings of prior GNN methods by approximating the convolution.

\item \textbf{SAGE}~\cite{hamilton2017inductive}. GraphSAGE focuses on inductive node classification but can also be applied for transductive settings.

\item \textbf{ChebNet}~\cite{defferrard2016convolutional}. ChebNet is a formulation of CNNs concerning spectral graph theory, which provides the necessary mathematical background and efficient numerical schemes to design fast localized convolutional filters on graphs.

\item \textbf{SGC}~\cite{wu2019simplifying}. Simple Graph Convolution~(SGC) reduces the excess complexity of GCNs by repeatedly removing the nonlinearities between GCN layers and collapsing the resulting function into a single linear transformation.

\item \textbf{APPNP}~\cite{klicpera2018predict}. Approximate personalized propagation of neural predictions~(APPNP) is a fast approximation to personalized propagation of neural predictions (PPNP), which utilizes this propagation procedure to construct a simple model. APPNP leverages a large, adjustable neighborhood for classification and can be easily combined with any neural network.
\end{itemize}

\section{Metrics of community detection}\label{sec:cdm}
The metrics are implemented by \url{https://networkx.org}. The \emph{modularity} is defined as $Q = \sum_{c=1}^{n}
\left[ \frac{L_c}{m} - \gamma\left( \frac{k_c}{2m} \right) ^2 \right]$, where the sum iterates over all communities $c$, $m$ is the number of edges, 
$L_c$ is the number of intra-community links for community $c$, 
$k_c$ is the sum of degrees of the nodes in community $c$, and $\gamma$ is the resolution parameter; The \emph{coverage} of a partition is the ratio of the number of intra-community edges to the total number of edges; The \emph{performance} of a partition is the number of intra-community edges plus inter-community non-edges divided by the total number of potential edges.

\end{document}